\documentclass[9pt,journal]{IEEEtran}
\pdfoutput=1

\usepackage{epsfig}
\usepackage{graphicx}
\usepackage{amsmath}
\usepackage{amssymb}

\usepackage[font=small,labelfont=bf]{caption}
\usepackage{makecell}

\usepackage{multirow}
\usepackage{tabularx}
\usepackage{dsfont}
\usepackage{url}
\usepackage[tight]{subfigure}
\usepackage{xcolor}
\usepackage{subfigure}
\usepackage{amsthm}
\usepackage[export]{adjustbox}
\usepackage{comment}

\usepackage[ruled,noend]{algorithm2e}
\usepackage[utf8]{inputenc} 
\usepackage[T1]{fontenc}    
\usepackage{url}            
\usepackage{booktabs}       
\usepackage{amsfonts}       
\usepackage{nicefrac}       
\usepackage{microtype}      

\usepackage{balance}

\def\argmax{\operatornamewithlimits{arg\,max}}

\def\PSNRL{$\mathrm{PSNR}_L$}
\def\SSIM{SSIM}
\def\BRIS{BRISQUE}

\definecolor{dg}{rgb}{0,0.694,0.298}
\definecolor{purple}{rgb}{0.4,0.176,0.569}
\definecolor{coralred}{RGB}{255,127,0}
\definecolor{caribbeangreen}{rgb}{0.0, 0.8, 0.6}
\definecolor{royalblue}{RGB}{65,105,225}

\usepackage{pifont}

\newcommand{\first}[1]{\textbf{\textcolor{red}{#1}}}
\newcommand{\second}[1]{\textbf{\textcolor{caribbeangreen}{#1}}}

\usepackage{colortbl}
\definecolor{tabgray}{rgb}{0.85,0.85,0.85}
\definecolor{top1}{rgb}{1.0, 0.6, 0.6} 
\definecolor{top2}{rgb}{0.98, 0.91, 0.71}
\definecolor{top3}{rgb}{0.91, 1.0, 1.0}
\definecolor{top1-2}{rgb}{1.0, 0.66, 0.66} 
\definecolor{top1-3}{rgb}{1.0, 0.72, 0.72} 
\definecolor{top1-4}{rgb}{1.0, 0.78, 0.78} 
\definecolor{top1-5}{rgb}{1.0, 0.84, 0.84} 
\definecolor{top1-6}{rgb}{1.0, 0.90, 0.90} 
\definecolor{top1-7}{rgb}{1.0, 0.96, 0.96}

\usepackage{xspace}
\makeatletter
\DeclareRobustCommand\onedot{\futurelet\@let@token\@onedot}
\def\@onedot{\ifx\@let@token.\else.\null\fi\xspace}
\def\eg{\emph{e.g}\onedot} 
\def\ie{\emph{i.e}\onedot} 
 
\def\etc{\emph{etc}\onedot} 
 
\def\etal{\emph{et al}\onedot}
\makeatother

\usepackage[colorlinks,
linkcolor=red,
citecolor=royalblue
]{hyperref}

\usepackage{scalerel}
\usepackage{tikz}
\usetikzlibrary{svg.path}
\definecolor{orcidlogocol}{HTML}{A6CE39}
\tikzset{
  orcidlogo/.pic={
    \fill[orcidlogocol] svg{M256,128c0,70.7-57.3,128-128,128C57.3,256,0,198.7,0,128C0,57.3,57.3,0,128,0C198.7,0,256,57.3,256,128z};
    \fill[white] svg{M86.3,186.2H70.9V79.1h15.4v48.4V186.2z}
                 svg{M108.9,79.1h41.6c39.6,0,57,28.3,57,53.6c0,27.5-21.5,53.6-56.8,53.6h-41.8V79.1z M124.3,172.4h24.5c34.9,0,42.9-26.5,42.9-39.7c0-21.5-13.7-39.7-43.7-39.7h-23.7V172.4z}
                 svg{M88.7,56.8c0,5.5-4.5,10.1-10.1,10.1c-5.6,0-10.1-4.6-10.1-10.1c0-5.6,4.5-10.1,10.1-10.1C84.2,46.7,88.7,51.3,88.7,56.8z};
  }
}
\newcommand\orcidicon[1]{\href{https://orcid.org/#1}{\mbox{\scalerel*{
\begin{tikzpicture}[yscale=-1,transform shape]
\pic{orcidlogo};
\end{tikzpicture}
}{|}}}}

\begin{document}
%
\title{\emph{Pasadena}: Perceptually Aware and Stealthy Adversarial Denoise Attack} 

\author{
\IEEEauthorblockN{Yupeng~Cheng\IEEEauthorrefmark{1}\,\orcidicon{0000-0003-3865-2947}\,,~\IEEEmembership{Student~Member,~IEEE}, Qing~Guo\IEEEauthorrefmark{1}\IEEEauthorrefmark{2}\,\orcidicon{0000-0003-0974-9299}\,,~\IEEEmembership{Member,~IEEE}, Felix~Juefei-Xu\,\orcidicon{0000-0002-0857-8611}\,,~\IEEEmembership{Member,~IEEE},  Shang-Wei~Lin\,\orcidicon{0000-0002-9726-3434}\,,~\IEEEmembership{Member,~IEEE},
Wei~Feng\,\orcidicon{0000-0003-3809-1086}\,,~\IEEEmembership{Member,~IEEE},
Weisi~Lin\,\orcidicon{0000-0001-9866-1947}\,,~\IEEEmembership{Fellow,~IEEE}, Yang~Liu\,\orcidicon{0000-0001-7300-9215}\,,~\IEEEmembership{Senior~Member,~IEEE}}
\thanks{
This work is supported in part by the Natural Science Foundation of Tianjin under Grant No. 20JCQNJC00720, the National Research Foundation, Singapore under its the AI Singapore Programme (AISG2-RP-2020-019), the National Research Foundation, Prime Ministers Office, 
Singapore under its National Cybersecurity R\&D Program (No. NRF2018NCR-NCR005-0001), NRF Investigatorship NRFI06-2020-0001, 
the National Research Foundation through its National Satellite of Excellence in Trustworthy Software Systems (NSOE-TSS) project under the National Cybersecurity R\&D (NCR)
Grant (No.~NRF2018NCR-NSOE003-0001), 
the Ministry of Education, Singapore, under its Academic Tier-2 Research Fund (Grant No. MOE2018-T2-1-068).

Yupeng Cheng, Qing Guo$^\dag$ (corresponding author. E-mail: tsingqguo@ieee.org), Shang-Wei Lin, Weisi Lin, and Yang Liu are with Nanyang Technological University, Singapore. Felix Juefei-Xu is with Alibaba Group, USA. Qing Guo and Wei Feng are with the School of Computer Science and Technology, College of Intelligence and Computing, Tianjin Key Laboratory of Cognitive Computing and Application, Tianjin University, Tianjin 300305, China, and are also with the Key Research Center for Surface Monitoring and Analysis of Cultural Relics (SMARC), State Administration of Cultural Heritage, China. 
Yupeng Cheng$^{*}$ and Qing Guo$^{*}$ are co-first authors.
}
}
\maketitle

\begin{abstract}
Image denoising can remove natural noise that widely exists in images captured by multimedia devices due to low-quality imaging sensors, unstable image transmission processes, or low light conditions.
Recent works also find that image denoising benefits the high-level vision tasks, \eg, image classification.
In this work, we try to challenge this common sense and explore a totally new problem, \ie, whether the image denoising can be given the capability of fooling the state-of-the-art deep neural networks (DNNs) while enhancing the image quality.
To this end, we initiate the very first attempt to study this problem from the perspective of adversarial attack and propose the \textit{adversarial denoise attack}.
More specifically, our main contributions are three-fold: \textit{First}, we identify a new task that stealthily embeds attacks inside the image denoising module widely deployed in multimedia devices as an image post-processing operation to simultaneously enhance the visual image quality and fool DNNs.
\textit{Second}, we formulate this new task as a kernel prediction problem for image filtering and propose the \textit{adversarial-denoising kernel prediction} that can produce adversarial-noiseless kernels for effective denoising and adversarial attacking simultaneously.
\textit{Third}, we implement an adaptive \textit{perceptual region localization} to identify semantic-related vulnerability regions with which the attack can be more effective while not doing too much harm to the denoising.
We name the proposed method as \textit{Pasadena} (Perceptually Aware and Stealthy Adversarial DENoise Attack) and validate our method on the NeurIPS'17 adversarial competition dataset, CVPR2021-AIC-VI: unrestricted adversarial attacks on ImageNet, and Tiny-ImageNet-C dataset. The comprehensive evaluation and analysis demonstrate that our method not only realizes denoising but also achieves a significantly higher success rate and transferability over state-of-the-art attacks.
\end{abstract}

\IEEEpeerreviewmaketitle

\section{Introduction}\label{sec:intro}
\begin{figure}[t]
    \centering
    \subfigure { 
    \includegraphics[width=0.95\linewidth]{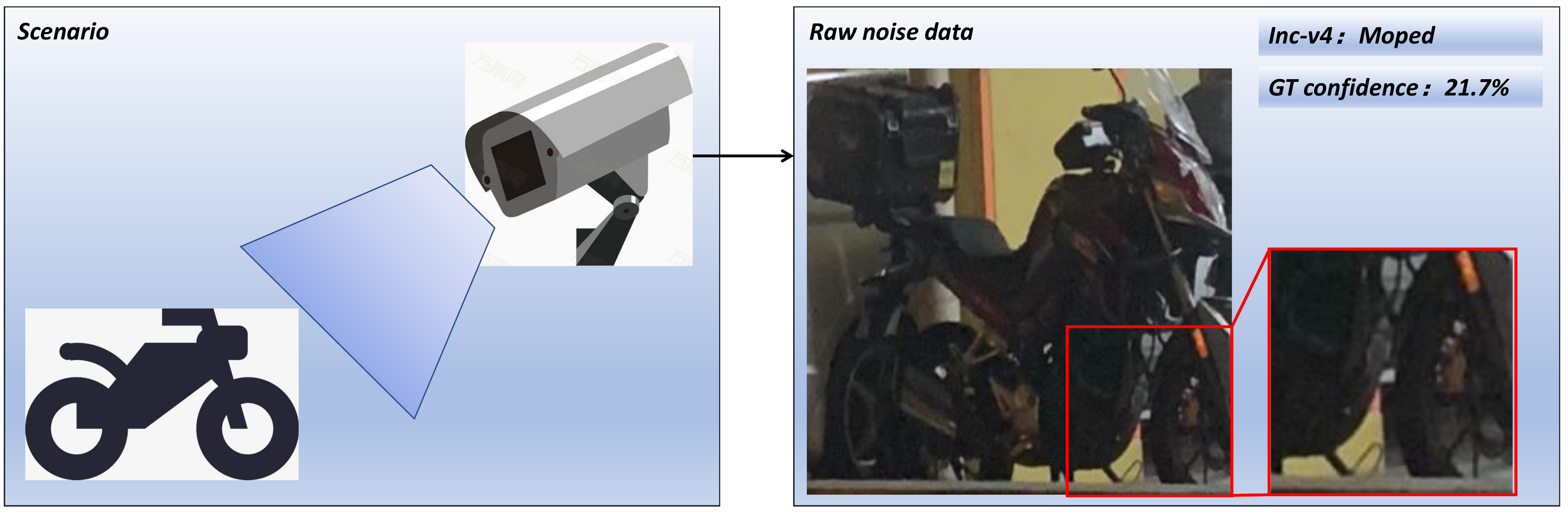}
    }
    \subfigure { 
    \includegraphics[width=0.95\linewidth]{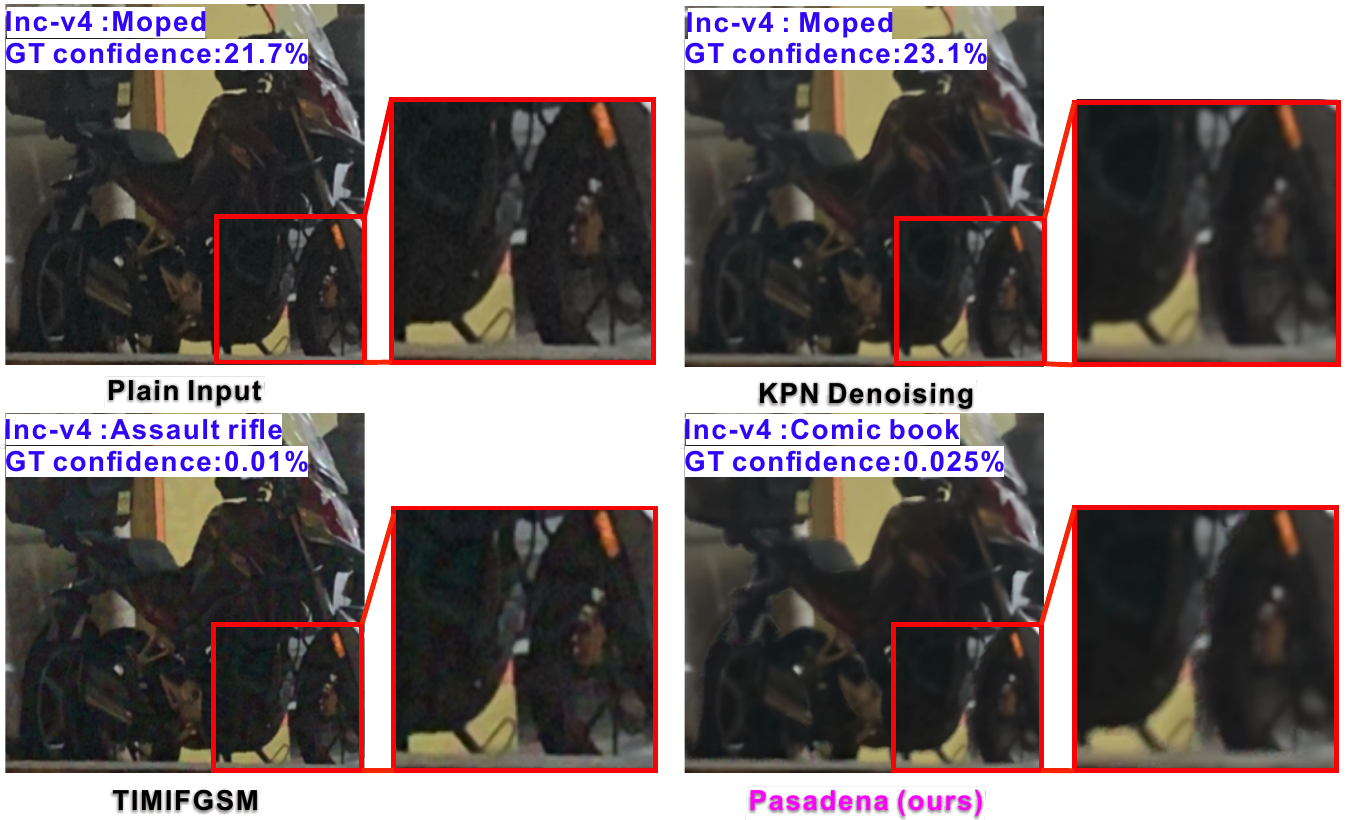}
    }
    \caption{Recognition and visualization results of a real noisy input image  (\ie, a \textbf{Moped} captured by iPhone XS at $8$ pm), its denoised counterpart, adversarial noise attacked version, and our adversarial denoise attacked result. The prediction results of Inc-v4 are displayed at the top left of each image, followed by the ground truth (GT) prediction confidence.
    The results show that the DNN, \eg, Inc-v4, is robust to natural noise, and correctly predicts the object's category even with the noisy input. The state-of-the-art (SOTA) denoising method, \eg, kernel prediction network (KPN) \cite{Mildenhall2018CVPR}, can improve the image quality and further enhance DNN's classification accuracy. The SOTA adversarial attack, \eg, TIMIFGSM \cite{dong2019evading}, misleads the DNN while corrupting the image severely. In contrast, our method \textit{Pasadena} can not only mislead the DNN but also significantly remove the noise pattern and enhance the image quality.} \label{fig:motivation}
\end{figure}

Image denoising is a fundamental computer vision problem. Its main objective is to remove natural noise that widely exists in the captured images due to low-quality imaging sensors, unstable image transmission processes, or low light conditions \cite{tian2020deep,cho2017geodesic,malladi2020image,ghosh2019fast}.
In recent years, with the rapid development of advanced deep learning techniques, the state-of-the-art deep neural networks (DNNs) have achieved significantly high or even near-human performance on high-level vision tasks, \eg, image classification \cite{he2016deep,dong2017cunet,zhang2019unsupervised} and detection \cite{guo2019distributed, cong2018hscs}.
In common sense, image denoising removing the noise corruption is able to benefit the high-level tasks. 
This has been validated and studied by performing denoising and high-level tasks jointly \cite{Liu2018IJCAI}. 
Specifically, as shown in Fig.~\ref{fig:motivation}, the powerful DNN, \ie, Inception-v4 (Inc-v4) \cite{Szegedy17Incv4}, predicts the real-world-captured noisy image as `moped' with $21.7\%$ confidence that is increased to $23.1\%$ with a deep learning-based denoiser for pre-processing \cite{krasin2017openimages}. 
%
%
In this work, we identify a new task to challenge this common understanding by giving the image denoising of the capability of fooling the state-of-the-art deep neural networks (DNNs), that is, we aim to remove the noise of input image while letting the DNNs predict incorrect category.
As shown in Fig.~\ref{fig:motivation}, we can enhance the real noisy image more clear but make Inc-v4 output `Comic book' instead of `Moped'.
In terms of the potential applications, this new task can not only pose the threat of image denoising to image classification but also has the potential to avoid maliciously recognition-based data collection or manipulation.
For example, DeepFake can automatically recognize an interested object (\eg,  the high-profile personnel's face) and switch with a maliciously generated fake one to influence the outcomes of various critical events \cite{DFDC2020,ijcai20_fakespotter}. 
Actually, the adversarial attack adding adversarial noise to the image is a straightforward solution to prevent such an abuse of recognition techniques and protect the interested object (\ie, the moped in Fig.~\ref{fig:motivation}) from being misused. Unfortunately, such a process will inevitably further corrupt the image and fail to provide visually clean perceptions, even rendering the generated image unacceptable for human beings. For example, the state-of-the-art attack, \ie, TIMIFGSM \cite{dong2019evading}, fools the advanced DNN, \ie, Inc-v4, while leaving traces of significantly severe noise patterns. 
%
%

In this paper, we initiate the first step to study the threat of image denoising to visual recognition from the perceptive of adversarial attack \cite{goodfellow2014explaining,moosavi2016deepfool,su2019one, guo2020spark,neurips20_abba,acmmm20_amora,tian2021ava} and propose the \textit{adversarial denoise attack} that aims to simultaneously denoise the input images while fooling the DNNs, which is actually a totally novel task that aims to stealthily embed the attacks inside the denoising module and enhances the image quality while making it hard to be recognized or analyzed by DNNs.
Note that, all existing image denoising methods cannot address this task and restore clean images mainly relying on fixed hand-crafted spatial filters, such as mean filters (arithmetic mean, geometric mean, harmonic mean, \etc.), median filter, min and max filters, \etc., or learning-based filters \cite{Bako2017TOG} and \cite{Mildenhall2018CVPR}. 

To realize an effective \textit{adversarial denoise attack}, we formulate it as a kernel prediction problem and propose the \textit{adversarial-denoising kernel prediction} producing adversarial-noiseless kernels for effective denoising and adversarial attacking simultaneously. Furthermore, we implement an adaptive \textit{perceptual region localization} to identify the semantic-related vulnerability regions with which the attack can be more effective while not doing too much harm to the denoising. Thus, our proposed method is termed \textit{Pasadena} (Perceptually Aware and Stealthy Adversarial DENoise Attack). As illustrated in Fig.~\ref{fig:motivation}, (top left) what is shown here is that the state-of-the-art (SOTA) DNN, \eg, Inc-v4, is robust to natural noise and usually can predict an object's category correctly even with noisy input. (Top right) the SOTA denoising method, \eg, kernel prediction network (KPN) \cite{Mildenhall2018CVPR}, can improve image quality and usually enhance the classification accuracy. (Bottom left) the SOTA adversarial attack, \eg, TIMIFGSM \cite{dong2019evading}, misleads the DNN while corrupting the image severely. (Bottom right) our proposed method, \textit{Pasadena}, can not only mislead the DNN but also at the same time improve the image quality.

We conduct comprehensive evaluation and analysis on the NeurIPS'17 adversarial competition dataset and demonstrate that our method not only performs effective denoising but also achieves a significantly higher success rate and transferability over the state-of-the-art attacks. 
In terms of the technical novelty in the field of adversarial attack, to the best of our knowledge, our method is the very first work trying to address the attacking problem from the view of denoising which is a widely used technique to enhance image quality. 
Our main contributions are summarized as follows:
\begin{itemize}
    \item We post a brand new problem, \ie, how to simultaneously remove the noise of the input images while fooling the DNNs.
    \item We identify this problem from the view of the adversarial attack and propose the very novel \textit{adversarial denoise attack} aiming to stealthily embed attacks inside the image denoising module.
    \item We formulate the adversarial denoise attack as a kernel prediction problem for image filtering and propose the \textit{adversarial-denoising kernel prediction} that can produce adversarial-noiseless kernels for effective denoising and adversarial attacking simultaneously.
    \item We implement an adaptive \textit{perceptual region localization} to identify the semantic-related vulnerability regions with which the attack can be more effective while not doing too much harm to the denoising.
    \item We comprehensively evaluate and analyze our method on the NeurIPS'17 adversarial competition, CVPR2021-AIC-VI: unrestricted adversarial attacks on ImageNet, and Tiny-ImageNet-C, and demonstrate that our method not only realizes effective denoising but also achieves a significantly higher success rate and transferability over the state-of-the-art attacks.
\end{itemize}


\section{Related Work}\label{sec:related}

\subsection{Image Denoising}
\label{subsec:related_denoising}
Noise commonly exists in images due to various sources such as the low-quality imaging sensor and unstable transmission, which affects many vision-based tasks, \eg, visual enhancement, feature extraction, and face recognition \cite{rahman2007video,luisier2010sure,jiang2016srlsp,ding2015robust,ghosh2019fast}, and makes image denoising a fundamental and important problem within the fields of signal processing and computer vision.
Many classic methods address image denoising as a statistics problem using analytical priors \cite{perona1990scale,rudin1992nonlinear,chen2011adaptive}. BM3D \cite{dabov2007image}, one of the most widely used algorithms tries to estimate the true signal by collaboratively filtering several similar image fragments and enhancing their sparsity in the frequency domain. 
Similarly, Guo \etal \cite{guo2015efficient} use the low-rank approximation to estimate and depress the noise in patches.
Based on the same assumption that similar noisy patches can be averaged to better estimate the true signal, multi-image denoising techniques including video or burst images have been built, such as VBM4D \cite{maggioni2011video}. It aligns similar image patches and jointly filters them by robust averaging. 

More recently, researchers utilize the power of DNNs to reach higher image quality \cite{heide2014flexisp, heide2016proximal, Bako2017TOG, Mildenhall2018CVPR,cho2018gradient}.
Among these methods, the kernel-prediction-based methods \cite{Bako2017TOG, Mildenhall2018CVPR} use a DNN to predict pixel-wise kernels and restore the noisy input by processing each pixel with its exclusive kernel. These methods have drawn great attention due to their high denoising performance.
More specifically, Bako \etal \cite{Bako2017TOG} denoise the Monte Carlo renderings using two pre-trained networks to predict the per-pixel filtering kernels. Inspired by this work, Mildenhall \etal \cite{Mildenhall2018CVPR} construct a UNet-based network to predict kernels for handling burst images and achieve impressive denoising performance.
Note that, although the advanced denoising methods can enhance the image quality significantly, they cannot let the denoised image fool DNNs thus is not suitable for the new task, \ie, adversarial denoise attack, which is the main objective of this paper.

\subsection{Adversarial Attacks}
%
Deep learning techniques have achieved great progress \cite{szegedy2016rethinking,huang2017densely,he2016deep} and benefited almost all kinds of computer vision tasks, \eg, image classification \cite{he2016deep,dong2017cunet,zhang2019unsupervised}, detection \cite{He_2017_ICCV}, segmentation \cite{ChenPKMY18}, denoising \cite{zhang2017beyond,chen2016deep,cho2018gradient}, deblurring \cite{su2017deep,liang2020raw}, demosaicking \cite{gharbi2016deep}, super-resolution \cite{tao2017detail}, and safety \& security-critical applications \cite{ackerman2017drive,rao2018deep,najafabadi2015deep,papernot2016towards,hinton2012deep}.
Nevertheless, recent works, \eg, \cite{szegedy2013intriguing,goodfellow2014explaining}, also discover a potential risk of using DNNs, \ie, a carefully crafted input named as the adversarial example can mislead a well-trained deep model and let it generate the wrong prediction with high confidence. The method for producing the adversarial examples is termed an adversarial attack.

Since then, numerous additive-perturbation-based adversarial attacks \cite{goodfellow2014explaining,moosavi2016deepfool,su2019one, guo2020spark} have been proposed with the main objectives of improving the high attack success rate and transferability across models while keeping the attacked image imperceptible for human beings. 
%
%
%
%
%
For example, Goodfellow \etal \cite{goodfellow2014explaining} implement a one-step attack method, \ie, the fast gradient sign method (FGSM), which is efficient but hard to achieve a high attack success rate.
Then, FGSM is further improved by using the iterative optimization \cite{BIM_2016_ICLRW} and the momentum term \cite{dong2018boosting}. 
More recently, \cite{dong2019evading} further explores how to improve the transferability of popular attacks, \eg, FGSM \cite{goodfellow2014explaining}, MIFGSM \cite{dong2018boosting}, and DIM \cite{xie2019improving}. 
%
Other kinds of attacks, \eg, DeepFool \cite{moosavi2016deepfool}, 
computes adversarial examples with tiny distortion in a simple and accurate manner regardless of time consumption.
Su \etal \cite{su2019one} indicate that merely one pixel's modification can totally confuse the neural networks classifier.
Papernot \etal \cite{Papernot2016TheLO} implement the Jacobian-based saliency map attack with an impressively high success rate, and Carlini \etal \cite{CW_2017_SSP} realize attacks with significantly imperceptible perturbations by optimizing elaborately designed objective functions under different norms.
Although achieving a great attack success rate, all existing attacks inevitably corrupt the input image and can lead to visually noisy perceptions for high transferability across DNNs. For example, the advanced TIMIFGSM \cite{dong2019evading} can mislead the Inc-v4 but introduce severe noise patterns.
This paper proposes a totally different and novel adversarial attack where the kernel prediction works as a way of embedding perturbations instead of using additive operation. Moreover, the kernels have the capability of both denoising and adversarial attack. As a result, our method can accomplish this very new task, \ie, adversarial denoise attack.
%

\begin{figure*}[]
    \centering
    \includegraphics[width=0.91\linewidth]{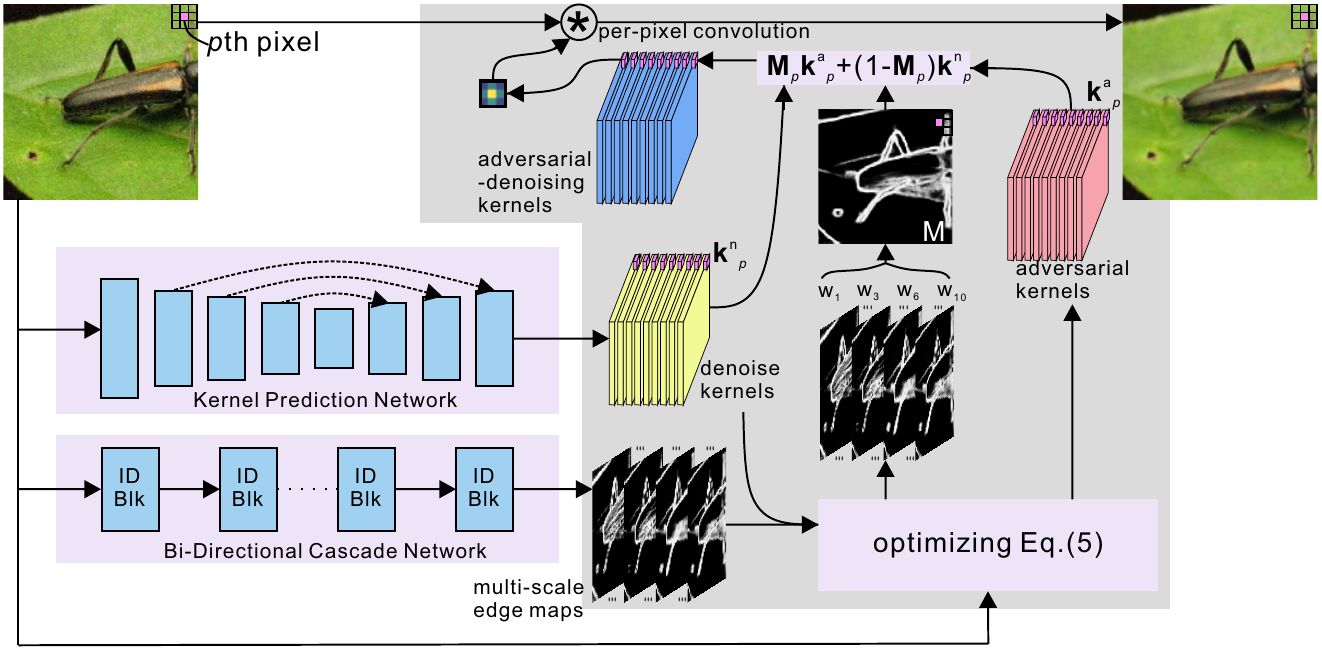}
    \caption{Pipeline of our Pasadena for adversarial denoise attack. \textit{First}, we propose the \textit{adversarial-denoising kernel prediction} where the adversarial kernels are predicted by optimizing the proposed adversarial objective function and the denoise kernels are calculated by employing a pre-trained kernel prediction network (KPN) \cite{Mildenhall2018CVPR}. \textit{Second}, we propose the \textit{perceptual region localization} where the semantic-related vulnerability regions are extracted by fusing multi-scale edge maps estimated by pre-trained bi-directional cascade network (BDCN) \cite{He2019CVPR}.
    }
\label{fig:pipeline}
\end{figure*}

\section{Methodology}\label{sec:method}

\subsection{Problem Formulation}
\label{subsec:formulation}
Given a noisy image $\mathbf{X}^{\mathrm{n}}$, we aim to produce an adversarial-noiseless image, \ie, $\mathbf{X}^{\mathrm{a}}$, that can fool a deep model while retaining higher quality than $\mathbf{X}^{\mathrm{n}}$. We name this task \textit{the adversarial denoise attack}.

We first review recent works on kernel-prediction-based image processing \cite{Mildenhall2018CVPR,Bako2017TOG,fu2021auto,guo2021efficientderain}. The main process can be summarized as follows:
\begin{align}\label{eq:kernel_den}
\mathbf{X}^\mathrm{a}_p=\mathrm{g}(\mathbf{X}^{\mathrm{n}}_p,\mathbf{k}_p)=\sum_{q\in\mathcal{N}(p)}\mathbf{X}^{\mathrm{n}}_qk_{pq},
\end{align}
where $p$ denotes the $p$-th noisy pixel in $\mathbf{X}^{\mathrm{n}}$, $\mathcal{N}(p)$ is the set of $p$'s neighbor pixels and has size of $N$, \ie, $p$ has $N$ neighboring pixels including itself. Then, we can process the noisy pixel by linearly combining its neighboring pixels where the combination weights are determined by a kernel, \ie, $\mathbf{k}_p=\{k_{pq}|q\in\mathcal{N}(p)\}$. Different pixels in $\mathbf{X}^{\mathrm{n}}$ can have different kernels 
and we denote all the kernels as $\mathcal{K}=\{\mathbf{k}_p\}$. Intuitively, the kernels determine the way of handling a noisy image. For example, when we let the values of $\mathbf{k}_p$ follow a Gaussian distribution that has the highest value at $k_{pp}$, Eq.~\eqref{eq:kernel_den} becomes a Gaussian denoising method. More recent works \cite{Mildenhall2018CVPR,Bako2017TOG} train a convolutional neural network (CNN) offline to predict a kernel for each noisy pixel and realize much better denoising results.

To realize the \textit{adversarial denoise attack}, we decompose $\mathcal{K}=\{\mathbf{k}_p\}$ into two parts, \ie, the adversarial kernel $\mathcal{K}^{\mathrm{a}}=\{\mathbf{k}^{\mathrm{a}}_p\}$ and the denoise kernel $\mathcal{K}^{\mathrm{n}}=\{\mathbf{k}^{\mathrm{n}}_p\}$. The first kernel aims to generate an adversarial example that can fool deep models while the second one is to improve the noisy image's quality. We then reformulate Eq.~\eqref{eq:kernel_den} as
\begin{align}\label{eq:kernel_adv}
\mathbf{X}^\mathrm{a}_p&=\mathrm{g}(\mathbf{X}^{\mathrm{n}}_p,\mathbf{M}_p,\mathbf{k}^\mathrm{a}_p,\mathbf{k}^\mathrm{n}_p) \nonumber\\
&=\sum_{q\in\mathcal{N}(p)}\mathbf{X}^{\mathrm{n}}_q(\mathbf{M}_pk^\mathrm{a}_{pq}+(1-\mathbf{M}_p)k^\mathrm{n}_{pq}),
\end{align}
where $\mathbf{M}$ is a perceptually-aware weight map with its values ranging from 0 to 1 and $\mathbf{M}_p$ denotes the $p$-th pixel of $\mathbf{M}$. Intuitively, $\mathbf{M}_p$ indicates if the $p$-th pixel should be denoised or attacked. For example, if $\mathbf{M}_p=1$, $\mathbf{X}^\mathrm{a}_p=\sum_{q\in\mathcal{N}(p)}\mathbf{X}^{\mathrm{n}}_qk^{a}_{pq}$, it means that $\mathbf{X}^\mathrm{a}_p$ is set to fool deep models, otherwise, it is a denoised pixel. 
Note that, we employ the kernel prediction method to achieve the desired adversarial denoising attack due to the following reasons: \ding{182} from the perspective of adversarial attack, kernel prediction method processes each pixel via an exclusive kernel (\ie, Eq.~\eqref{eq:kernel_den}), allowing each pixel to be tuned independently according to the adversarial objective function as done in the additive-noise-based attacks. As a result, we are able to realize a high attacking success rate and transferability. 
\ding{183} In terms of denoising, the kernel prediction method is a state-of-the-art method \cite{Mildenhall2018CVPR,tian2020deep}. In general, the kernel prediction method is fed with the noisy image and predicts suitable kernels to filter each pixel, showing better image restoration capability than transitional methods \cite{dabov2007image,ghosh2017artifact,ghosh2017pruned,ghosh2019fast}.
In particular, BM3D \cite{dabov2007image} utilizes correlations between image patches and employs collaborative filtering for denoising. S. Ghosh \etal \cite{ghosh2017pruned,ghosh2017artifact,ghosh2019fast} estimate kernels for filtering via heuristic ways. In contrast, the kernel prediction method takes the advantage of deep learning and predicts kernels according to the embedding of the input image. As a result, it can generalize to different image contents.

When using Eq.~\eqref{eq:kernel_adv} to produce the noiseless but adversarial examples, we should consider the following questions: 1) Given a noisy image, how to estimate the adversarial kernel, \ie, $\mathcal{K}^{\mathrm{a}}$, and the denoise kernel, \ie, $\mathcal{K}^{\mathrm{n}}$, effectively? 2) How to estimate the perceptually aware weight map, \ie, $\mathbf{M}$, which should be sparse (\ie, to make sure that most of the image regions are denoised) and perceptual (\ie, to make sure semantic-dependent regions are attacked). We will detail the solutions for the two questions in Section~\ref{subsec:ad_kernel} and \ref{subsec:region_loc} and summarize the attack algorithm in Section~\ref{subsec:attack}.

\subsection{Adversarial-Denoising Kernel Prediction}
\label{subsec:ad_kernel}
Given a DNN for image classification denoted as $\phi(\cdot)$ and 
a noisy image $\mathbf{X}^\mathrm{n}$, we predict the classification label $y$ of the image via $\phi(\mathbf{X}^\mathrm{n})$.
Our attack method is to generate an adversarial-noiseless example, \ie, $\mathbf{X}^{\mathrm{a}}$, which can let the DNN predict an incorrect label. 

\textit{First}, to denoise the noisy image, we propose to estimate the denoise kernel, \ie, $\mathcal{K}^\mathrm{n}=\{\mathbf{k}_p^\mathrm{n}\}$, through the recent kernel-prediction-based denoising method \cite{Mildenhall2018CVPR}. It takes noisy images as input and trains a pretrained CNN to predict spatially varying denoise kernels that can remove a wide range of noises.

\textit{Second}, to realize effective attack, we build the following objective function and optimize it to get the required adversarial kernels, \ie, $\mathcal{K}^\mathrm{a}=\{\mathbf{k}_p^\mathrm{a}\}$, as well as the weight map $\mathbf{M}=\{\mathbf{M}_p\}$:
\begin{align}\label{eq:add_adv_obj}
\argmax_{\mathbf{M},\mathcal{K}^\mathrm{a}} \, &\lambda J_1(\phi(\{\mathrm{g}(\mathbf{X}^{\mathrm{n}}_p,\mathbf{M}_p,\mathbf{k}^\mathrm{a}_p,\mathbf{k}^\mathrm{n}_p)\}),y)-\gamma J_2(\mathcal{K}^\mathrm{a},\mathcal{K}^\mathrm{n})\nonumber\\
&\mathrm{~~~~subject~to~~~~} \forall p, \|\mathbf{k}_p^\mathrm{a}\|_{0}\leq\epsilon, 
\end{align}
where $J_1(\cdot)$ denotes the crossing entropy loss for the objective of generating adversarial kernels, \ie, fooling the deep model $\phi(\cdot)$, and $J_2(\cdot)$ is the $L_2$ norm loss that encourages the adversarial kernels $\{\mathbf{k}^\mathrm{a}_p\}$ to be similar to the denoise kernels $\{\mathbf{k}^\mathrm{n}_p\}$ in order to produce a high quality image. The parameters $\lambda$ and $\gamma$ control the balance between the two loss functions. The `$\{\cdot\}$' denotes the set of pixel-depended variables. For example, $\{\mathrm{g}(\mathbf{X}^{\mathrm{n}}_p,\mathbf{M}_p,\mathbf{k}^\mathrm{a}_p,\mathbf{k}^\mathrm{n}_p)\}$ is the set of $\mathbf{X}^\mathrm{a}$'s pixels. Each pixel of $\mathbf{X}^\mathrm{n}$ has an adversarial kernel, \eg, $\mathbf{k}_p^\mathrm{a}$, and a weight, \eg, $\mathbf{M}_p$.

The constrain term, \ie, $\|\mathbf{k}_p^\mathrm{a}\|_{0}\leq\epsilon$, requires that the size of $\mathbf{k}_p^\mathrm{a}$ should not be larger than $\epsilon$. An oversized kernel makes the attack succeed easily but heavily corrupts the original noisy image with worse quality.

We can calculate the gradient of the objective functions with respect to both the adversarial kernels and the weight map, thus realizing the gradient-based attack. We will show that such an attack can also achieve impressively high transferability. However, it should be noted that the weight map is only tuned by the objective of maximizing the loss function, which helps achieve a high success rate but harms the effectiveness of denoising. The desired strategy is to only attack the semantic-related regions while denoising other regions to guarantee that the image quality is further improved.

\subsection{Perceptual Region Localization}
\label{subsec:region_loc}

As introduced in Sec.~\ref{subsec:formulation}, the weight map $\mathbf{M}$ should be both sparse and perceptual to realize effective \textit{adversarial-denoise attack}. However, tuning the weight map $\mathbf{M}=\{\mathbf{M}_p\}$ by solving Eq.~\eqref{eq:add_adv_obj} directly cannot achieve these goals. In this section, we propose a perceptual region locator to produce sparse and perceptual $\mathbf{M}$. 

We first use a state-of-the-art perceptual edge detector, \ie, bi-directional cascade network \cite{He2019CVPR}, to extract multi-scale edge maps of the input noisy image, \ie, $\{\mathbf{E}^i\}_{i=1}^L=\varphi(\mathbf{X}^\mathrm{n})$ where $\varphi(\cdot)$ is the bi-directional cascade network and $\{\mathbf{E}^i\}_{i=1}^L$ denote edge detection results of $L$ scales. Then, we get $\mathbf{M}=\{\mathbf{M}_p\}$ by combining these edge maps via
\begin{align}\label{eq:bdcn}
\mathbf{M}_p=\mathrm{Sigmoid}\left(\sum_{i=1}^{L}w_i\mathbf{E}^i_p - \theta\right) 
\end{align}
where $\mathbf{E}^i_p$ and $\mathbf{M}_p$ are the $p$-th pixel of $\mathbf{E}^i$ and $\mathbf{M}$, respectively. $\mathrm{Sigmoid}(\cdot)$ is a general activation function mapping the output to [0, 1]. $\theta$ is the bias for potential threshold. 
The intuitive motivation behind the idea is that the multi-scale edges are sparse (\ie, most of the pixels of each edge map are labeled as zero, \ie, non-edge) and perceptual (\ie, covering the main semantic information). As a result, their linear combination, \ie, our desired weight map, is naturally inherited to these properties. 
Then, we can reformulate the
objective function in Eq.~\eqref{eq:add_adv_obj} as 
\begin{align}\label{eq:add_adv_obj2}
\argmax_{\{\mathbf{w}_i\},\mathcal{K}^\mathrm{a}} &\, \lambda J_1(\phi(\{\mathrm{g}(\mathbf{X}^{\mathrm{n}}_p,\sum_{i=1}^{L}w_i\mathbf{E}^i_p,\mathbf{k}^\mathrm{a}_p,\mathbf{k}^\mathrm{n}_p)\}),y)-\gamma J_2(\mathcal{K}^\mathrm{a},\mathcal{K}^\mathrm{n})\nonumber\\
&\mathrm{~~~~subject~to~~~~} \forall p, \|\mathbf{k}_p^\mathrm{a}\|_{0}\leq\epsilon.
\end{align}

\subsection{Attack Algorithm}
\label{subsec:attack}
With the proposed \textit{adversarial-denoising kernel prediction} and \textit{perceptual region localization}, we then realize the desired \textit{adversarial denoise attack} and show the whole pipeline in Fig.~\ref{fig:pipeline}. More specifically, we summarize the whole process of our attack method in Algorithm~1. Given a noisy image $\mathbf{X}^\mathrm{n}$, we first estimate the denoise kernels via a kernel-prediction network (KPN) \cite{Mildenhall2018CVPR}, and the multi-scale edge maps via the bi-directional cascade network (BDCN) \cite{He2019CVPR}. 
More specifically, we use the implementation \cite{Mildenhall2018CVPR} that adopts the encoder-decoder architecture with skip connections for KPN whose parameters are pre-trained \cite{Mildenhall2018CVPR} on a synthetic training dataset \cite{krasin2017openimages}.
We follow \cite{He2019CVPR} for the realization of BDCN that is trained on multi-scale ground truth edge maps to encourage the learning of multi-scale representations at different layers.
Then, we calculate the adversarial kernels and perceptually-aware weight maps by optimizing Eq.~\eqref{eq:add_adv_obj2} with the hyper-parameters, \ie, the kernel size $\epsilon=5$, loss weight $\lambda=1-\gamma=0.9$, and step size $\alpha=0.1$. After $T=10$ iterations, we finally obtain the adversarial-noiseless image, \ie, $\mathbf{X}^\mathrm{a}$, via Eq.~\eqref{eq:kernel_adv}.
Among the hyper-parameters, the loss of weight plays a key role in the balance between the attack success rate and the image quality and we will discuss the influence of the $\lambda$ in the experimental section \ref{subsubsec:comp_baselines_wa_quatitative}.

\begin{algorithm}[tb]
	\label{alg}
	{
		\caption{\fontfamily{bch}\selectfont \footnotesize{Pasadena for Adversarial Denoise Attack}}
		\KwIn{A noisy image $\mathbf{X}^\mathrm{n}$; the pre-trained kernel-prediction network (KPN); the pre-trained bi-directional cascade network (BDCN), \ie, $\varphi(\cdot)$; the attacked model $\phi(\cdot)$; hyper-parameters: $\epsilon=5$, $\lambda=1-\gamma=0.9$, the threshold $\theta=0.45$, step size $\alpha=0.1$, and iteration number $T=10$.}
		\KwOut{Adversarial-noiseless image, \ie, $\mathbf{X}^\mathrm{a}$.}
		Predict the denoise kernels $\mathcal{K}^\mathrm{n}=\mathrm{kpn}(\mathbf{X}^\mathrm{n})$\;
		Estimate the multi-scale edge maps $\{\mathbf{E}^i\}_{i=1}^L=\varphi(\mathbf{X}^\mathrm{n})$ \;
		Randomly initialize the $\{w_i\}$ in the range [0,1] and $\sum_i{w_i}=1$\;
		Initialize the $\mathcal{K}^\mathrm{a}=\mathcal{K}^\mathrm{n}$\;
		Obtain the ground truth label by $y=\phi(\mathbf{X}^\mathrm{n})$\;
 		\For{$t=1\ \mathrm{to}\ T$}{
            Calculate the gradient of the objective functions in Eq.~\eqref{eq:add_adv_obj2} w.r.t. $\{w_i^{t}\}$ and $\mathcal{K}^{\mathrm{a},t}$ and get $\{\nabla{w_i^t}\}$ $\{\nabla_{\mathbf{k}_p^{\mathrm{a},t}}\}$\;
            Update $\{w_i\}$ and $\mathcal{K}^\mathrm{a}$ by $\{w_i^{t+1}=w_i^{t}+\alpha \nabla{w_i^{t}}\}$ and $\{\mathbf{k}_p^{\mathrm{a},t+1}=\mathbf{k}_p^{\mathrm{a},t}+\alpha \nabla{\mathbf{k}_p^{\mathrm{a},t}}\}$  \;
            Calculate the perceptual weight map by $\{\mathbf{M}_p^{t+1}=            \mathrm{Sigmoid}\left(\sum_{i=1}^{N}w_i^{t+1}\mathbf{E}^i_p - \theta\right)\}$\; 
            Calculate the adversarial-noiseless image by $\{\mathbf{X}^\mathrm{a}_p =\mathrm{g}(\mathbf{X}^{\mathrm{n}}_p,\mathbf{M}_p^{t+1},\mathbf{k}^{\mathrm{a},t+1}_p,\mathbf{k}^\mathrm{n}_p)\}$\;
            \If{$\phi(\mathbf{X}^\mathrm{a})\neq y$}{
                return $\mathbf{X}^\mathrm{a}$ \;
            } 
    		$t=t+1$\;
    	}
	}
\end{algorithm}%

\begin{figure*}
    \centering
    \subfigure { 
    \includegraphics[width=0.31\linewidth]{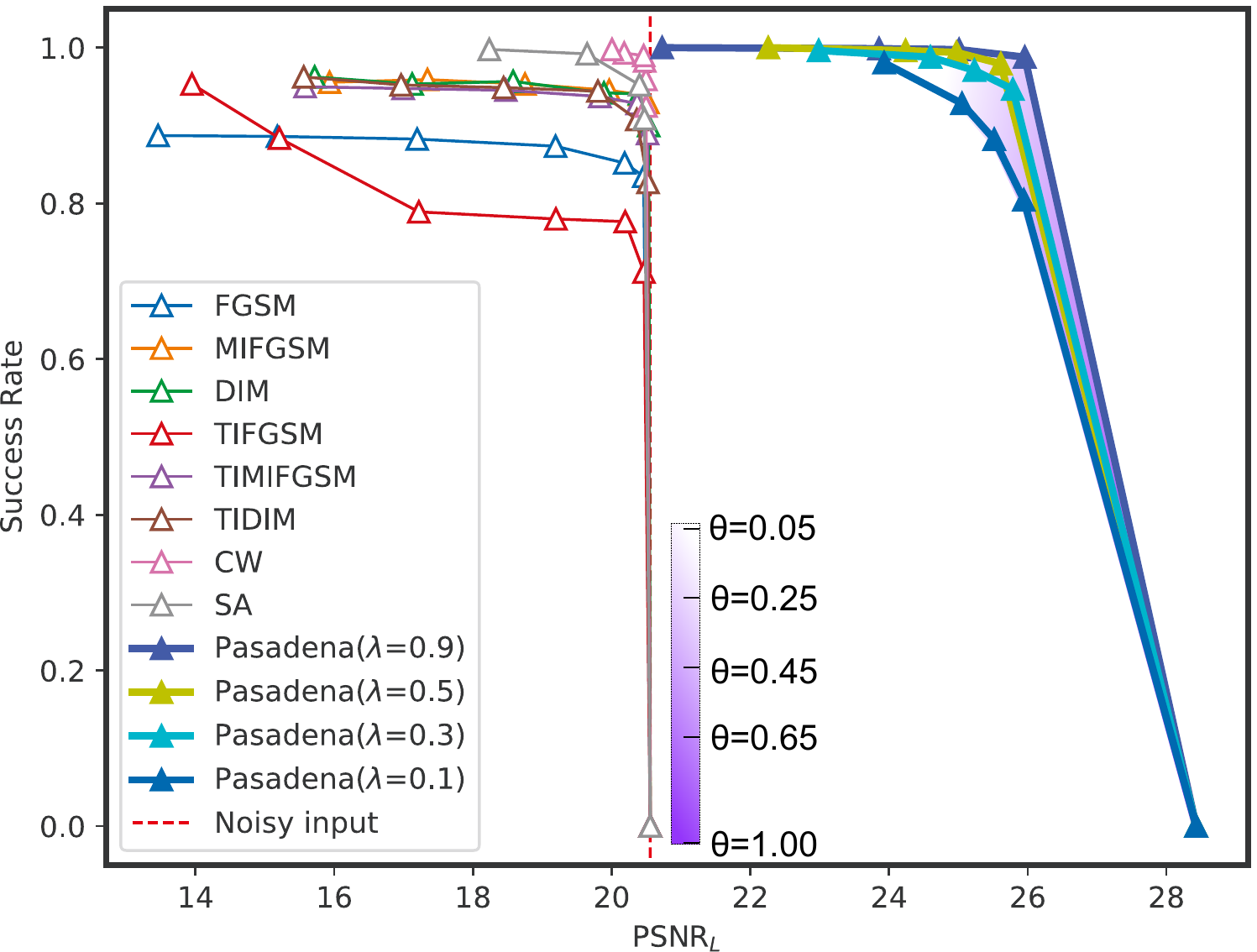}
    } 
    \subfigure { 
    \includegraphics[width=0.31\linewidth]{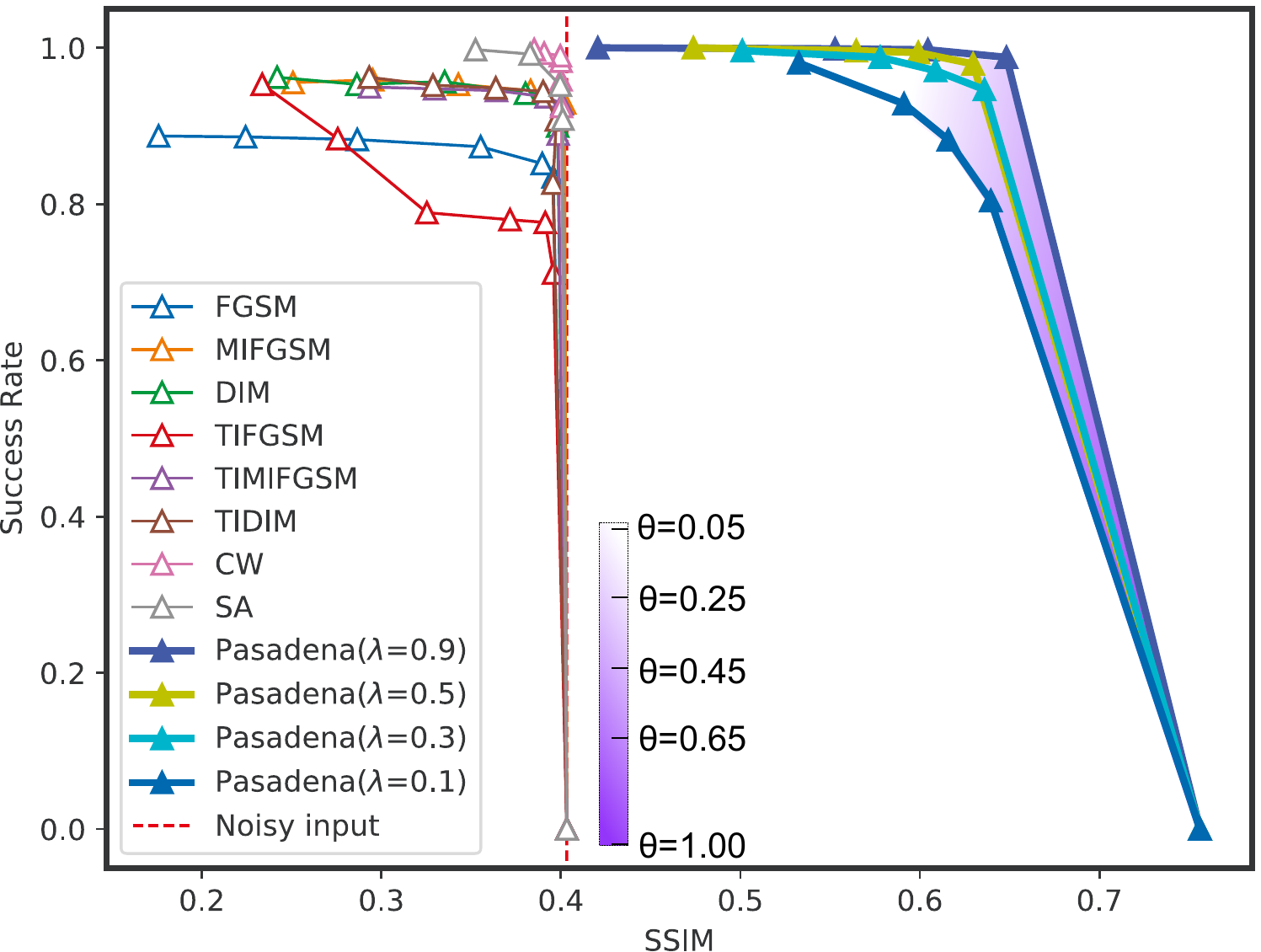}
    } 
    \subfigure { 
    \includegraphics[width=0.31\linewidth, height=42.5mm]{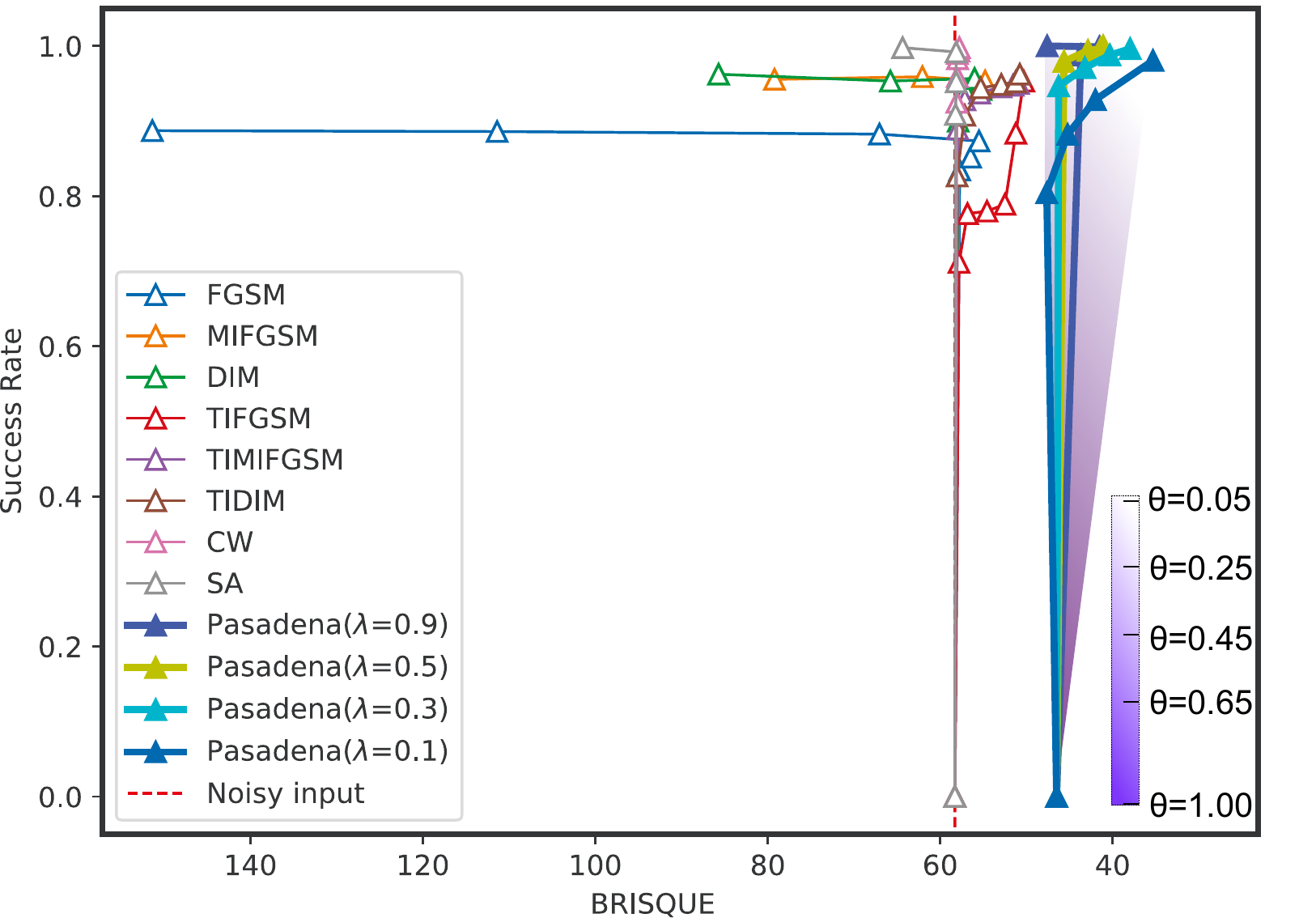}
    } 
    \caption{Success rates of whitebox attack along with $\mathrm{PSNR}_{L}$, SSIM, and BRISQUE for eight baseline methods with noise inputs and our four variants with different adversarial loss weights $\lambda$. The curves of Pasadena are generated by setting threshold $\theta$ ranging from $0.05$ to $0.65$. {For the six attacks, \ie, FGSM/MIFGSM/DIM/TIFGSM/TIMIFGSM/TIDIM, we tune the maximum perturbation ranging from $4$ to $55$ with the max intensity of 255. For C\&W attack, we tune the weight $c$ ranging from $0.01$ to $100$. For SA attack, we tune the $k$ ranging from $5000$ to $50000$.}
    The color gradient indicates the $\theta$ variation with the attack success rate.
    }
\label{fig:exp_noise}
\end{figure*}

\begin{figure*}[t]
    \centering
    \includegraphics[width=1.0\linewidth]{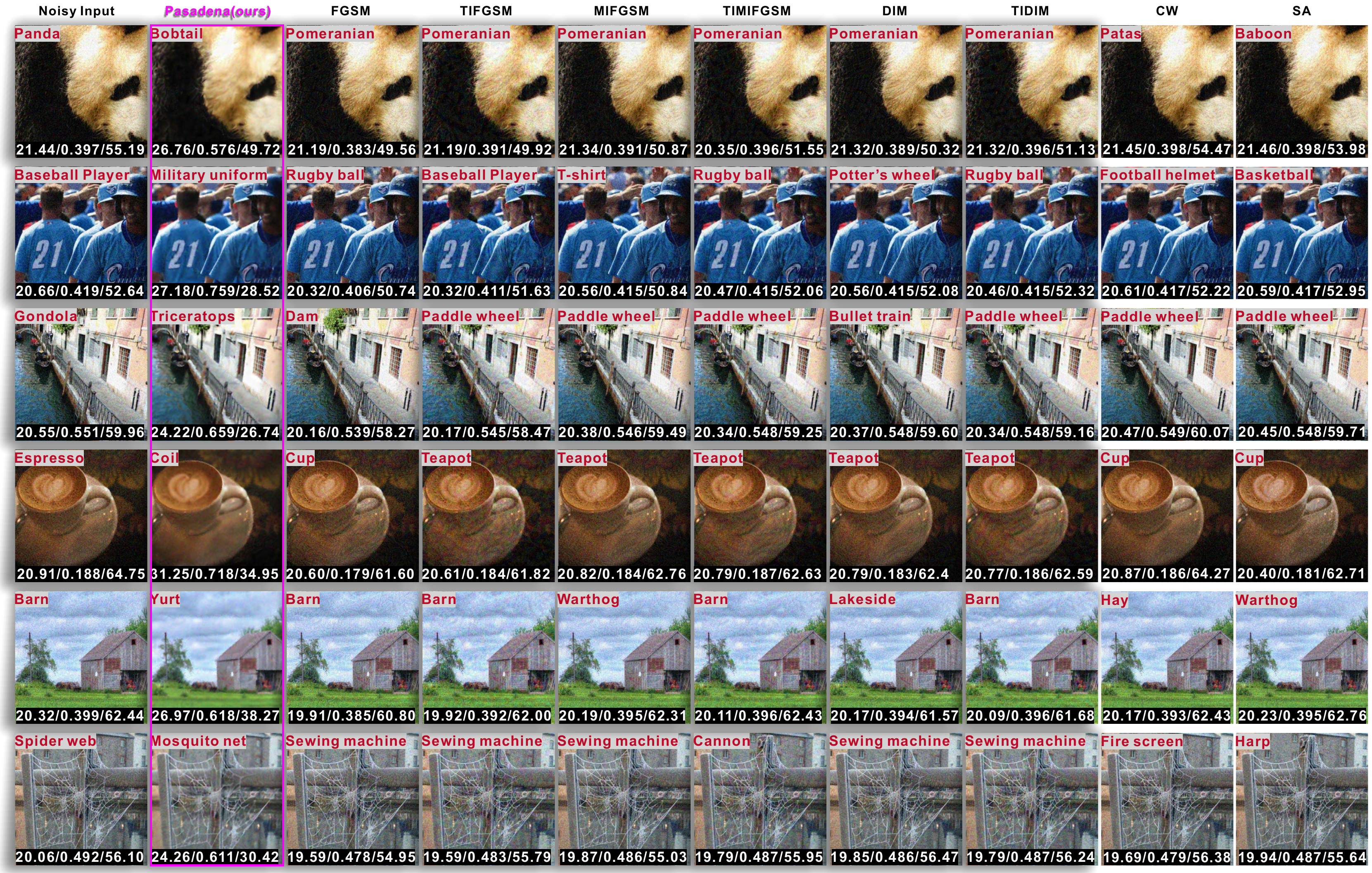}
    \caption{Visualization results of adversarial examples crafted for Inception-v4, using our method Pasadena and eight baseline attacks. For each image, its prediction of Inception-v4 is displayed on top-left. Three numbers at the bottom refer to $\mathrm{PSNR}_{L}$, SSIM and BRISQUE values. All noisy inputs are correctly classified to their ground truth label.}
\label{fig:via_baselines}
\end{figure*}

\section{Experiments}\label{sec:exp}
In this section, we illustrate our experimental results to demonstrate the capability of our framework in adversarial attacking while increasing the image quality. We first describe the experimental settings in Sec.~\ref{exp_setup}. Afterward, we compare the results of our framework with six state-of-the-art attacks in Sec.~\ref{subsec:comp_baselines_wa}. Then, we also compare our method with a very competitive solution for joint denoising and attack, which leads to another six attacks, in Sec.~\ref{subsec:comp_baselines_wadenoi}. At last, we perform the ablation study to validate the contribution of each component of our method in Sec.~\ref{subsec:Ablation_wa}.
\subsection{Setup}
\label{exp_setup}
\textbf{Dataset and Models:}
We conduct our experiments on NeurIPS'17 adversarial competition, \ie, DEV dataset \cite{kurakin2018adversarial}, CVPR2021-AIC-VI: unrestricted adversarial attacks on ImageNet \cite{cvpr2021-aic-vi, dong2020benchmarking}, and Tiny-ImageNet-C dataset \cite{hendrycks2019robustness}. The first two datasets are ImageNet-compatible. Specific to our task, we manually add Gaussian white noise ($\sigma=0.1$) to the clean images of DEV and CVPR2021-AIC-VI datasets to test the image quality increasing as well as the attack success of the adversarial attack methods. We use the Tiny-ImageNet-C dataset to validate the generalization of our method to shot noise and impulse noise.
To verify our performance in various networks, we introduce four models that are trained on the standard ImageNet dataset, including Inception v3 (Inc-v3) \cite{szegedy2016rethinking}, Inception v4 (Inc-v4), Inception ResNet v2 (IncRes-v2) \cite{szegedy2017inception}, and ResNet50 \cite{he2016deep}.
%

\textbf{Metrics. } 
To evaluate the effectiveness of the \textit{adversarial denoise attack}, which fools the classifier and denoises the image at the same time, we select the attack success rate and image quality for measuring the performance of the attack methods. There are two widely used full-reference image quality assessment metrics: peak signal-to-noise ratio (PSNR) and structural similarity (SSIM) \cite{wang2004image}:
\begin{align}\label{eq:PSNR}
\mathrm{PSNR}(\mathbf{X}, \mathbf{Y}) &= 20 \cdot \log_{10} \left( \frac{\mathrm{MAX}_\mathrm{I}}{\mathrm{MSE}(\mathbf{X},\mathbf{Y})} \right)\\
\mathrm{SSIM}(\mathbf{X}, \mathbf{Y}) &= \frac{(2\mu_x\mu_y+c_1)(2\sigma_{xy}+c_2)}{(\mu^2_x+\mu^2_y+c_1)(\sigma^2_x+\sigma^2_y+c_2)},
\end{align}
where $\mathbf{X}$ and $\mathbf{Y}$ are the two images being compared. $\mu_x$ and $\mu_y$ are the mean values of $\mathbf{X}$ and $\mathbf{Y}$ with corresponding variances $\sigma^2_x$ and $\sigma^2_y$. $\sigma_{xy}$ is the covariance. $\mathrm{MAX}_\mathrm{I}$ is the max intensity, \eg, $255$ under 8-bit representation.

PSNR measures the difference between two images across all pixels and ignores the local structural information, thus the measurement does not align very well with human perception. To alleviate this issue, we propose a variant of the PSNR denoted as $\mathrm{PSNR}_{L}$ to measure the difference between two images by considering their local patches 
\begin{align}\label{eq:LPSNR}
\mathrm{PSNR}_{L}(\mathbf{X},\mathbf{Y}) =  \cfrac {1} {S} \sum_{i=1}^S \mathrm{PSNR}(\mathbf{W}^\mathrm{X}_i, \mathbf{W}^\mathrm{Y}_i),
\end{align}
where $\mathbf{W}^\mathrm{X}_i$ and $\mathbf{W}^\mathrm{Y}_i$ are the $i$-th patches in the two images $\mathbf{X}$ and $\mathbf{Y}$, respectively. 
Since the PSNRs of all local patches are calculated independently, $\mathrm{PSNR}_{L}$ measures the global and local differences at the same time.
$S$ is the number of patches. During the implementation, we fix the patch size as $8\times8$ pixels and sample patches via a sliding window with the stride of $1$. 
Besides, we also choose a widely used non-reference image quality assessment metric, \ie, dubbed blind/referenceless image spatial quality evaluator (BRISQUE) \cite{mittal2012no}, to further evaluate the naturalness of images. 
This metric utilizes scene statistics of locally normalized luminances coefficients, named as mean subtracted contrast normalized (MSCN), to quantify possible losses of `naturalness' in the image. Specifically, given an image $\mathbf{X}$, the MSCN coefficient on pixel $\mathbf{X}(i,j)$ is formulated as:
\begin{align}\label{eq:MSCN}
\mathrm{MSCN}(i, j) =  \cfrac {\mathbf{X}(i,j)-\mu(i,j)} {\sigma(i,j)+C} ,
\end{align}
where $i \in 1,2...M, j \in 1,2...N$ are the indices, and $M$, $N$ are the image height and width. $C=1$ is a stability keeping constant parameter when the denominator tends to zero. $\mu(i,j)$ and $\sigma(i,j)$ are local weighted mean and local weighted standard deviation, respectively. Note that, different from $\mathrm{PSNR}_{L}$ or SSIM, the lower BRISQUE value means better image quality.
Finally, we select the attack success rate, $\mathrm{PSNR}_{L}$, SSIM, and BRISQUE for measuring the general performance of attack methods in our task.

\textbf{Attack Baselines.} 
First, we select six advanced adversarial attacks as the baseline methods: fast gradient sign method (FGSM) \cite{goodfellow2014explaining}, momentum iterative fast gradient sign method (MIFGSM) \cite{dong2018boosting}, diverse inputs method (DIM) \cite{xie2019improving}, as well as their translation-invariant version proposed in \cite{dong2019evading} which are denoted as TIFGSM, TIMIFGSM, and TIDIM, respectively. In particular, we set the Gaussian kernel (kernel size $= 15$ and $\sigma=3$) following Dong's report \cite{dong2019evading}. 
Furthermore, we also choose another two challenging attacks getting significantly imperceptible image perturbations, \ie, C\&W \cite{CW_2017_SSP} (with $L_2$ norm) and projected gradient descent (PGD) version of sparse attack (SA)~\cite{croce2019sparse}.

\textbf{Denoise~\&~Attack Baselines.}
We also compare our method with a very competitive solution for joint denoising and attack, \ie, first denoising the noisy inputs and then attacking the denoised images. As a result, we obtain another eight baseline methods (\ie, `Denoise+FGSM/MIFGSM/DIM/TIFGSM/TIMIFGSM/TIDIM/C\&W/SA') with the eight attacks and a state-of-the-art denoising method \cite{Mildenhall2018CVPR}.


\begin{table*}[t]
\centering
\small
\caption{Adversarial comparison results on DEV dataset whose images are added the additive Gaussian noise ($\sigma$=0.1). We use eight baselines and the proposed attacks against four normally trained models: ResNet50, Inc-v3, Inc-v4, and IncRes-v2, considering both white-box and black-box attacks for each model. The attacks contain six baseline methods, \ie, FGSM/MIFGSM/DIM/TIFGSM/TIMIFGSM/TIDIM, using the maximum perturbation of $8$, C\&W with $c=1.0$, SA with $k=10000$ and the proposed Pasadena. For each compared group (\ie, the four columns for ResNet50, Inc-v4, IncRes-v2, and Inc-v3), white-box attack results are shown in the 
first column. The rest three columns display the black-box attack results representing the transferability. 
We highlight the best two results with red and green fonts, respectively.
}
\label{Tab-noisecompare}
\setlength{\tabcolsep}{1mm}{}{
\begin{tabular}{l|cccc|cccc|cccc}

\toprule

\rowcolor{tabgray}\multicolumn{1}{c|}{Crafted from} & \multicolumn{4}{c|}{Inc-v3} & \multicolumn{4}{c|}{Inc-v4} & \multicolumn{4}{c}{IncRes-v2} \\

\cmidrule(r){1-1} \cmidrule(r){2-5} \cmidrule(r){6-9} \cmidrule(r){10-13}
\rowcolor{tabgray} Attacked model
& Inc-v3  &  Inc-v4   &   IncRes-v2 &  ResNet50

& Inc-v4  &  Inc-v3   &   IncRes-v2 &   ResNet50

& IncRes-v2 &  Inc-v3      &  Inc-v4 &    ResNet50\\

\midrule

FGSM 
&89.5 &33.0  &34.2    &44.0  
&85.2 &38.0  &40.2    &47.4 
&81.2 &39.8  &37.9    &47.9        \\

TIFGSM 
&83.1 & 26.6  &22.4  &45.0   
&77.7 & 23.7  &25.8  &46.0  
&73.4 & 25.9  &28.5  &48.2         \\

MIFGSM      
&93.2 &18.2  &19.8   &34.2   
&94.0 &32.4  &32.1   &39.8 
&96.8 &41.7  &36.7   &46.9          \\

TIMIFGSM      
&92.9 &23.0  &20.3  &41.4   
&92.8 &24.7  &27.0  &45.3
&93.8 &30.6  &35.3  &52.9        \\

DIM      
&94.3  &\second{48.3}  &\second{49.2}  &52.3   
&94.1  &\second{56.6}  &\second{57.7}  &\second{56.6} 
&95.1  &\second{63.1}  &\second{60.4}  &\second{61.9}        \\

TIDIM      
&91.7  &39.8  &35.0  &\second{54.6}    
&90.8  &35.4  &40.2  &55.7  
&89.6  &40.6  &47.1  &61.8         \\

C\&W ($L_2$ norm)      
&\second{99.5} &23.8	&21.7	&26.2 	
&\second{99.0} &26.7	&22.2	&27.5 	
&\second{98.6} &27.6	&24.9	&29.2         \\

SA      
&95.6 &10.8	&3.5	&25.1 	
&95.2 &10.6	&3.9	&25.7 	
&95.2 &9.9	&4.2	&26.1

        \\

Pasadena (ours) 
&\first{100.0}   &\first{75.8}  &\first{72.7} &\first{78.2}   
&\first{100.0}   &\first{72.7}   &\first{74.1} &\first{78.0}  
&\first{100.0}  &\first{78.4} &\first{81.3} &\first{81.0}           \\

\bottomrule

\end{tabular}
}
\end{table*}

\begin{table*}[t]
\centering
\small
\caption{Adversarial comparison results on CVPR2021-AIC-VI \cite{cvpr2021-aic-vi,dong2020benchmarking} whose images are added the additive Gaussian noise ($\sigma$=0.1). We use eight baselines and the proposed attacks against four normally trained models: ResNet50, Inc-v3, Inc-v4, and IncRes-v2, considering both white-box and black-box attacks for each model. The attacks contain six baseline methods, \ie, FGSM/MIFGSM/DIM/TIFGSM/TIMIFGSM/TIDIM, using the maximum perturbation of $8$, C\&W with $c=1.0$, SA with $k=10000$ and the proposed Pasadena. For each compared group (\ie, the four columns for ResNet50, Inc-v4, IncRes-v2, and Inc-v3), white-box attack results are shown in the first column. The rest three columns display the black-box attack results representing the transferability. 
We highlight the best two results with red and green fonts, respectively.
}
\label{Tab-devtianchi}
\setlength{\tabcolsep}{1mm}{}{
\begin{tabular}{l|cccc|cccc|cccc}

\toprule

\rowcolor{tabgray}\multicolumn{1}{c|}{Crafted from} & \multicolumn{4}{c|}{Inc-v3} & \multicolumn{4}{c|}{Inc-v4} & \multicolumn{4}{c}{IncRes-v2} \\

\cmidrule(r){1-1} \cmidrule(r){2-5} \cmidrule(r){6-9} \cmidrule(r){10-13}
\rowcolor{tabgray} Attacked model
& Inc-v3  &  Inc-v4   &   IncRes-v2 &  ResNet50

& Inc-v4  &  Inc-v3   &   IncRes-v2 & ResNet50 

& IncRes-v2  &  Inc-v3      &  Inc-v4 & ResNet50  \\

\midrule

FGSM 
&79.5	&\second{37.5}	&\second{38.0}	&32.8
&76.5   &\second{39.9}   &\second{38.5}   &\second{33.8}   
&69.0	&\second{38.4}	&\second{36.3}	&\second{33.0}           \\

TIFGSM 
&70.4	&30.5	&27.9	&\second{33.7}  
&69.4   &29.1   &28.3   &32.2   
&60.7	&28.8	&30.6	&32.0           \\

MIFGSM      
&\first{99.9}	&14.9	&13.8	&17.2
&\first{99.8}   &18.5   &14.8   &18.6  
&\second{99.3}	&19.0	&16.0	&18.6           \\

TIMIFGSM      
&98.5	&13.8	&12.3	&17.9
&98.4   &15.9   &12.9   &18.5   
&95.6	&17.4	&15.5	&19.9        \\

DIM      
&98.9	&15.4	&14.0	&23.5  
&97.3	&19.8	&15.9	&25.6   
&95.8	&21.4	&17.6	&26.6         \\

TIDIM      
&93.9	&16.1	&13.3	&26.9  
&93.5	&17.5	&14.1	&26.1   
&89.5	&18.9	&18.1	&28.8          \\

C\&W ($L_2$ norm)      
&99.4 &27.5	&28.5	&25.6	
& \second{98.6} &33.6	&30.7	&27.9	
&97.4 &35.0	&31.3	&28.3        \\

SA      
&97.6 &28.1	&26.9	&30.2	
&95.4 &32.2	&27.7	&30.9	
&95.3 &33.4	&30.5	&31.1

        \\

Pasadena (ours) 
&\second{99.7}  &\first{86.3} &\first{79.8} &\first{78.2}   
&\first{99.8}  &\first{79.7} &\first{78.0} &\first{76.7} 
&\first{99.6}  &\first{83.5} &\first{87.3} &\first{79.0}       \\

\bottomrule

\end{tabular}
}
\end{table*}

\subsection{Comparison with Attack Baselines}
\label{subsec:comp_baselines_wa}

\subsubsection{Quantitative Analysis}
\label{subsubsec:comp_baselines_wa_quatitative}
We first demonstrate the denoising and attack ability of our framework in this part by evaluating the performance of adversarial examples crafted for the Inc-v4 model on DEV dataset.
Note that, since this work focuses on both image quality and the ability of adversarial attack, we should compare all methods according to the quality, \eg, $\mathrm{PSNR}_L$, of generated adversarial examples and the attack success rate.
To this end, for each attack, we tune the success rate and the image quality-related parameters and get Succ.~Rate-$\mathrm{PSNR}_L$/SSIM/BRISQUE curves for clear visualization comparison (\eg, Fig.~\ref{fig:exp_noise}).
More specifically, for our method, we slide the threshold $\theta$ in Eq.~\eqref{eq:bdcn} from $0.05$ to $0.65$ to tune the success rate.
{
For the attacks including FGSM, MIFGSM, DIM, TIFGSM, TIMIFGSM, and TIDIM, we tune the maximum perturbation ranging from $4$ to $55$ with the max intensity of 255.
In terms of the C\&W attack, its attack strength is decided by the weight of the adversarial objective function denoted as $c$, and we tune it ranging from $0.01$ to $100$.}
{For the SA attack, the upper limit of the number of perturbed pixels (\ie, $k$) determines the attack strength and we set $k$'s range from $5000$ to $50000$.}
Furthermore, to evaluate the effect of loss weight (\ie, $\lambda$ in Eq.~\eqref{eq:add_adv_obj}), we construct four variants of our method by setting $\lambda=0.9, 0.5, 0.3, 0.1$. 

We show the comparison results on DEV dataset between our four variants and six baseline methods in  Fig.~\ref{fig:exp_noise}.  
Generally, all attack success rates have a negative correlation with the image quality (\ie, PSNR$_L$, SSIM, and BRISQUE) because heavily adversarial perturbation can effectively disturb the model prediction while reducing the image quality.
Moreover, we have the following observations:

\textit{First}, \textit{Pasadena} can accomplish the new task, \ie, \textit{adversarial denoise attack}, effectively with the capability of enhancing the image quality (\ie, higher \PSNRL{}, \SSIM{}, and lower \BRIS{} over the noisy input) and realizing significantly higher attack success rate.
In contrast, all of the additive-perturbation-based attacks further corrupt the noisy input, leading to lower \PSNRL{} and \SSIM{} as well as higher \BRIS.
It can be noted that the TIFGSM, TIMIFGSM, and TIDIM can clearly achieve better \BRIS{} values than noisy inputs when the success rate is low. 
The main reason is that \BRIS{} evaluates the image's naturalness based on the MSCN coefficients defined in Eq.~\eqref{eq:MSCN} that measure whether the image's local smoothness and contrast meet those of nature images.
As a result, the noise-like adversarial perturbations usually lead to high \BRIS{} and the three translation-invariant attacks break the noise perturbations pattern locally by using a Gaussian filter (\ie, Fig.~\ref{fig:exp_noise}), thus can achieve better (\ie, lower) \BRIS{} scores. 
However, as shown in Fig.~\ref{fig:exp_noise}, their noise pattern is not removed and leads to lower \PSNRL{} and \SSIM{}.


\textit{Second}, we further see that our method achieves a much higher attack success rate than baseline methods under similar image quality. 
{For example, the baseline methods with the maximum perturbation of $8$ (for FGSM, MIFGSM, DIM and there TI versions),  the C\&W wit $c=1.0$, the SA method with $k=10000$ and our method with $\theta=0.05, \lambda=0.9$ have similar \SSIM{} values of $0.4$. As shown in Table~\ref{Tab-noisecompare}, our method achieves the attack success rate of $100.0\%$, while the best baseline (\ie, DIM) has the success rate of $99\%$.}

%
%

\textit{Third}, in Fig.~\ref{fig:exp_noise}, the curves of our four variants with different $\lambda$ (\ie, the loss weight in Eq.~\eqref{eq:kernel_den} depicting different curves) and $\theta$ (\ie, the threshold in Eq.~\eqref{eq:bdcn} determines the different points on the same curves), show their great influence to the attacking and denoising performance. 
%
It is easy to see that lower thresholds (\ie, $\theta$) can lead to stronger attack, \ie, the attacking success rate of $\lambda=0.1$ (yellow curve) increases from $80.0\%$ to $99\%$ with the $\theta$ decreases from $0.65$ to $0.05$. 
In terms of the loss weight, a higher $\lambda$ results in more effective attack under the same image quality. For example, with the same $\epsilon = 0.05$, the attacking success rate of $\lambda = 0.1$ is $99\%$, which is slightly lower than $100\%$ of $\lambda = 0.9$. Nevertheless, the SSIM value $\lambda = 0.1$ is $0.54$ and much higher than $0.42$ for $\lambda = 0.9$.


\begin{figure*}
    \centering
    \subfigure { 
    \includegraphics[width=0.31\linewidth]{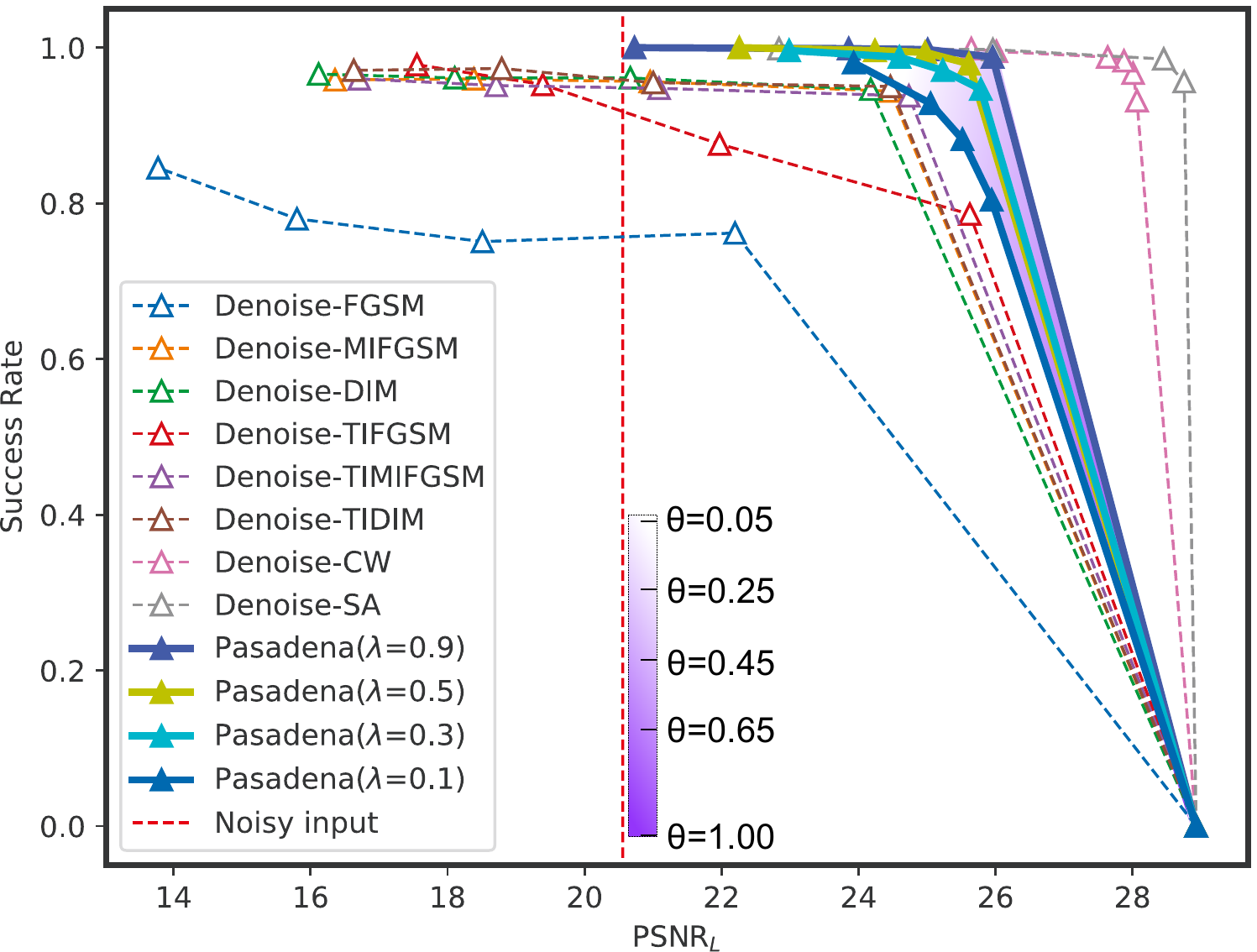}
    } 
    \subfigure { 
    \includegraphics[width=0.31\linewidth]{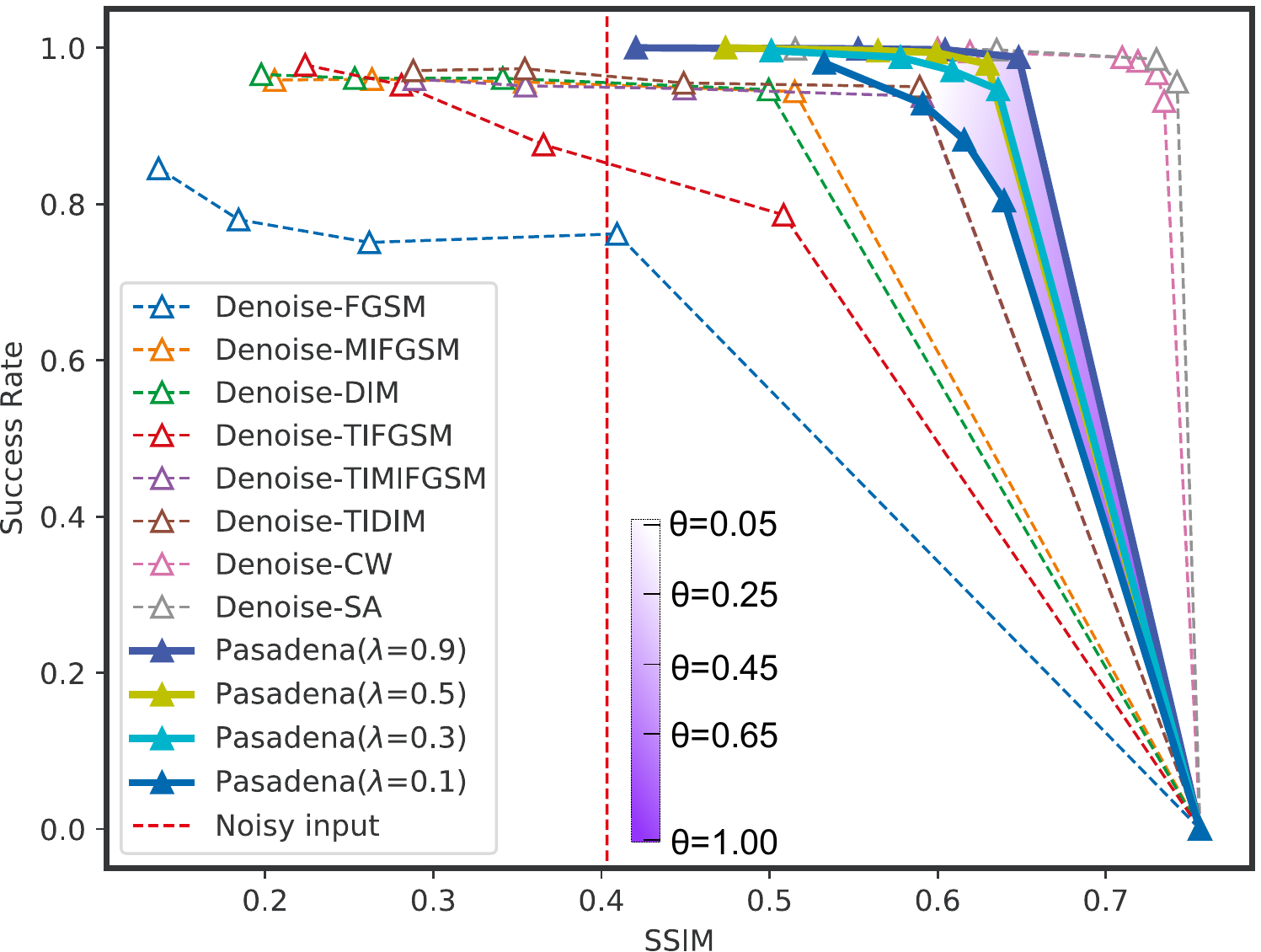}
    } 
    \subfigure { 
    \includegraphics[width=0.31\linewidth, height=42.5mm]{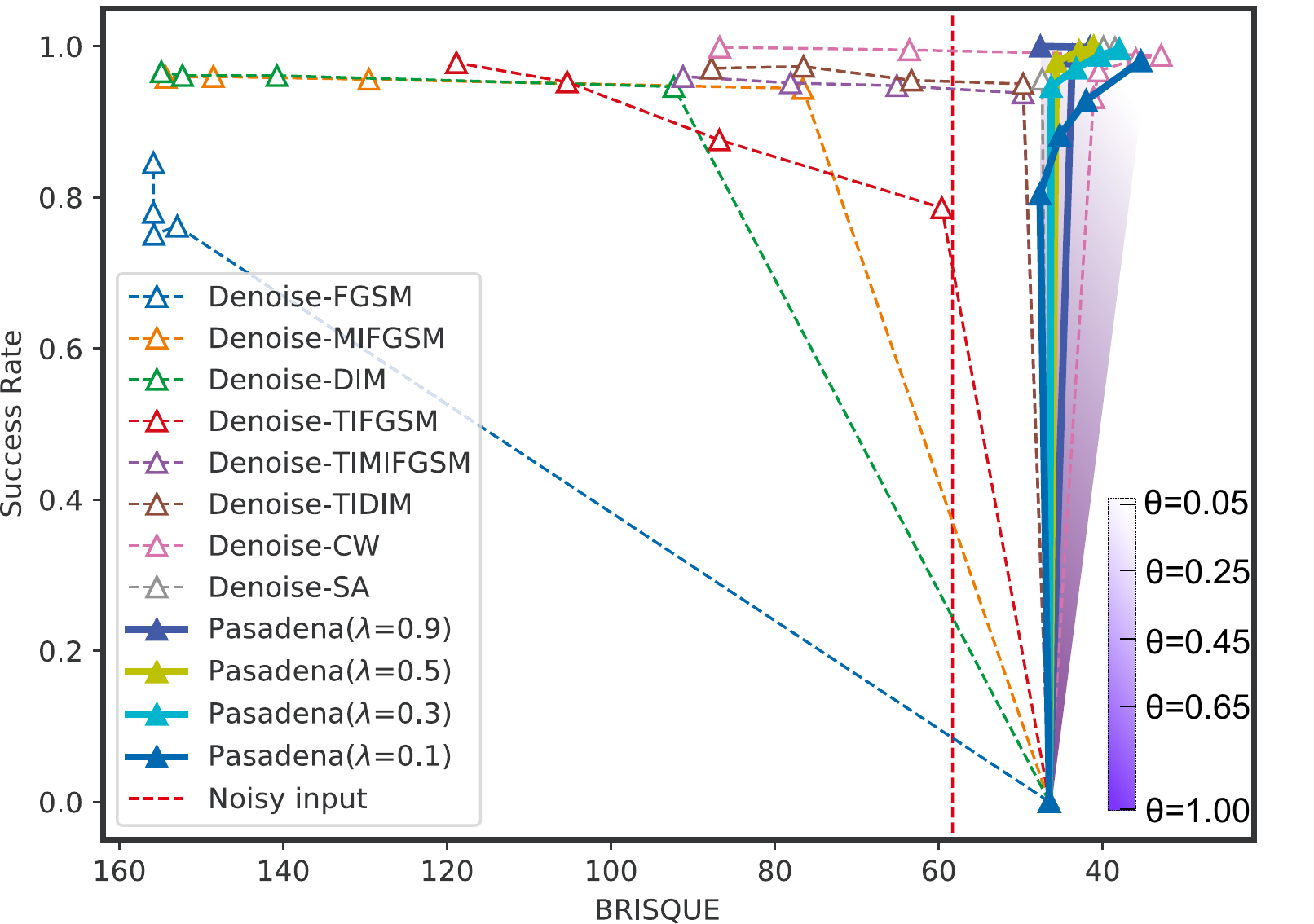}
    } 
    \caption{Success rates of whitebox attack along with $\mathrm{PSNR}_{L}$, SSIM, and BRISQUE for eight baseline methods with \textbf{DENOISED} inputs and our four variants with different adversarial loss weights $\lambda$. The curves of Pasadena are generated by setting threshold $\theta$ ranging from $0.05$ to $0.65$.
    {For the six attacks, \ie, Denoise+FGSM/MIFGSM/DIM/TIFGSM/TIMIFGSM/TIDIM, we tune the maximum perturbation ranging from $16$ to $55$ with the max intensity of 255. For Denoise+C\&W attack, we tune the weight $c$ of adversarial loss ranging from $0.01$ to $100$. For Denoise+SA attack, we tune the maximum perturbation pixels number $k$ ranging from $5000$ to $50000$.}
    The color gradient indicates the $\theta$ variation w.r.t. the attack success rate.
    }
\label{fig:exp_denoise}
\end{figure*}

\begin{figure*}[t]
    \centering
    \includegraphics[width=1.0\linewidth]{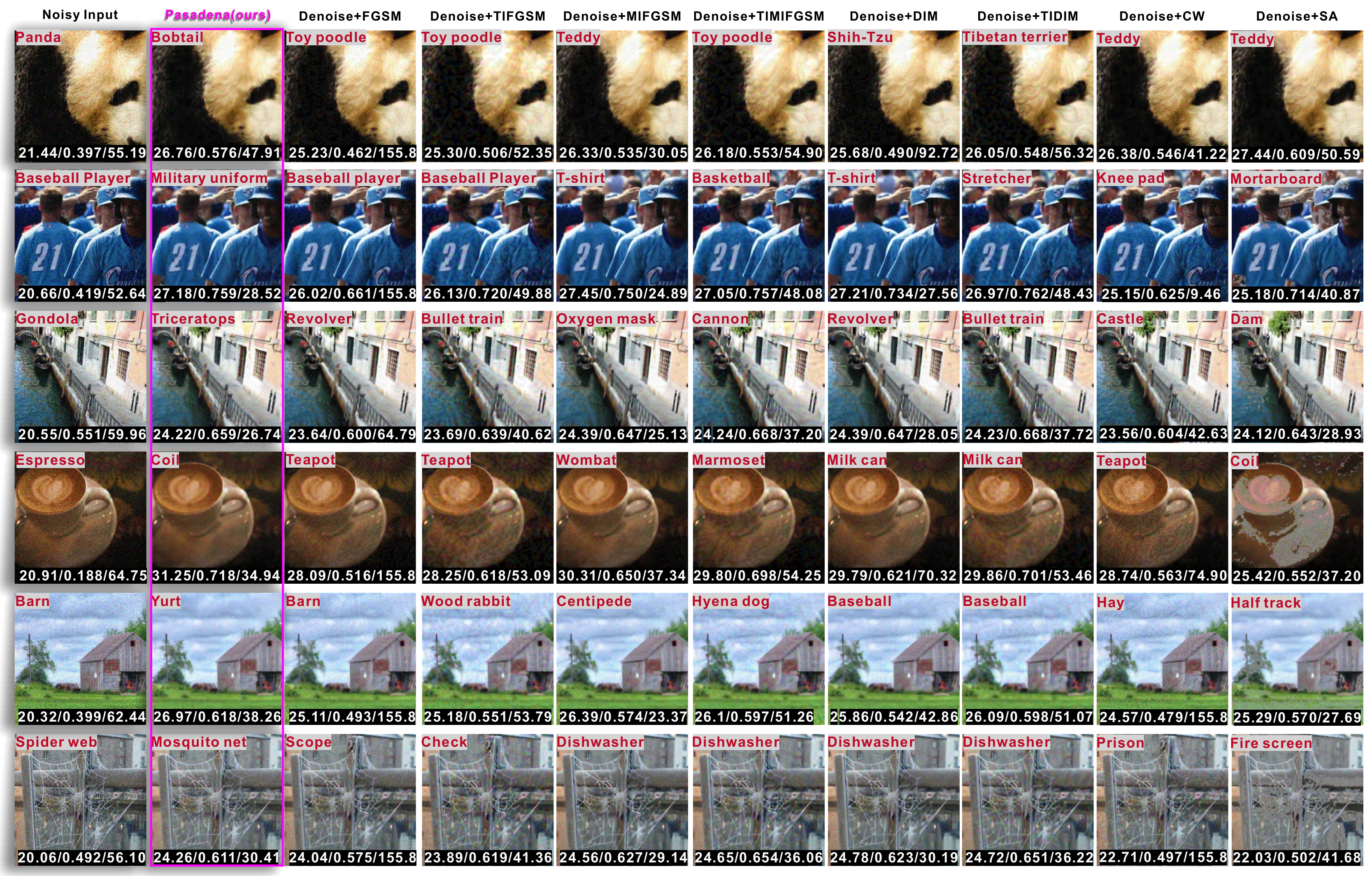}
    \caption{Visualization results of adversarial examples crafted for the Inception-v4, using our method Pasadena and eight baseline attacks. Different from Fig.~\ref{fig:via_baselines}, there is a denoising procedure (KPN) before baseline attacks. For each image, its prediction of the Inception-v4 is displayed on the top-left. Three numbers at the bottom refer to $\mathrm{PSNR}_{L}$, SSIM, and BRISQUE values. All noisy inputs are correctly classified to their ground truth label.}
\label{fig:via_baselines_denoise}
\end{figure*}

\begin{table*}[t]
\centering
\small
\caption{Adversarial comparison results on DEV dataset with additive Gaussian noise ($\sigma$=0.1). It contains the success rates (\%) of black \& white--box adversarial attacks among four normally trained models, ResNet50, Inc-v3, Inc-v4, and IncRes-v2. {The attacks contain six baseline methods (\ie, Denoise+FGSM/MIFGSM/DIM/TIFGSM/TIMIFGSM/TIDIM) with maximum perturbation of $8$, Denoise+C\&W with $c=10.0$, Denoise+SA with $k=30000$ and the Pasadena.} For each compared group (\ie, the four columns for ResNet50, Inc-v4, IncRes-v2, and Inc-v3), white-box attack results are shown in the first one. The rest three columns display the black-box attack results representing the transferability. 
We highlight the best two results with red and green fonts, respectively.
}
 
\label{Tab-denoisecompare}
\setlength{\tabcolsep}{1mm}{}{
\begin{tabular}{l|cccc|cccc|cccc}

\toprule

\rowcolor{tabgray}\multicolumn{1}{c|}{Crafted from} & \multicolumn{4}{c|}{Inc-v3} & \multicolumn{4}{c|}{Inc-v4} & \multicolumn{4}{c}{IncRes-v2} \\

\cmidrule(r){1-1} \cmidrule(r){2-5} \cmidrule(r){6-9} \cmidrule(r){10-13}
\rowcolor{tabgray} {Attacked model}
&  Inc-v3 &  Inc-v4   &   IncRes-v2 &  ResNet50

&  Inc-v4 &  Inc-v3   &   IncRes-v2 &  ResNet50

&   IncRes-v2 &  Inc-v3      &  Inc-v4 &  ResNet50\\

\midrule

Denoise+FGSM 
&78.8 &37.6 &40.3  &41.7  
&76.0 &41.9 &41.3  &39.0
&70.9 &45.9 &43.8  &43.6      \\

Denoise+TIFGSM 
&77.2 &42.1  &42.1  &49.8
&74.3 &39.7  &41.1  &45.4
&70.3 &43.2  &46.8  &31.1     \\

Denoise+MIFGSM      
&97.3   & 12.3  &14.1     &54.4
&98.3 &19.0   &21.8  &57.6
&\first{100.0}   &29.9 & 25.5       &66.7 
\\

Denoise+TIMIFGSM      
&96.6 &8.9  &6.1       &63.1
&97.4  &6.3   &11.0 &65.7
&96.4   &13.3 &23.6     &75.4  
\\

Denoise+DIM      
&97.3   &  36.2  &  36.7    &66.8 
&97.9  &  43.0   &  42.1 &69.7
&98.0   &  51.4 &  48.4    &79   
\\

Denoise+TIDIM      
&93.4  &24.4  &16.4      &\second{73.6}
&94.2  &16.6   &21.2  &\second{75.5}
&91.1  &21.6 &34.5   &\first{81.6}     
\\

Denoise+C\&W ($L_2$ norm)      
&\first{100.0}	&\second{46.5}	&\second{48.5}	&48.4 
&99.5   &\second{49.6}	&\second{51.3}	&51.5	
& \second{99.0}   &\second{52.1}	&\second{50.2}	&55.5 
        \\

Denoise+SA      
& \second{99.9}          &35.7	&33.7	&46.8
&\second{99.8}  &39.9	&33.0	&46.4	
&\first{100.0}  &43.9	&43.1	&49.6
        \\

Pasadena (ours)      
&\first{100.0} &\first{75.8}  &\first{72.7}  &\first{78.2} 
&\first{100.0} &\first{72.7}  &\first{74.1}  &\first{78.0}
&\first{100.0} &\first{78.4}  &\first{81.3}  &\second{81.0}      
\\

\bottomrule

\end{tabular}
}
\end{table*}

\subsubsection{Qualitative Analysis}
\label{subsubsec:comp_baselines_wa_qualitative}
We show six examples of DEV dataset in Fig.~\ref{fig:via_baselines} to compare the visualization results qualitatively.
Obviously, all attack baselines let the noisy input be worse with lower \PSNRL{} and \SSIM{} and produce a more salient noise pattern.
In contrast, our method can generate visually clean or denoised adversarial images with improved \PSNRL{}, \SSIM{} and \BRIS{}.
More specifically, TIMIFGSM perturbs the `Panda' image and successfully misleads the Inc-v4 model to predict it as `Pomeranian' but poses a terrible image quality.
While, our method removes the main noise while preserving the main edges and also misleading the DNN to produce the `Bobtail' prediction.
We can find similar results on other images.
%
%

\subsubsection{Comparison on Transferability}
\label{subsec:comp_baselines_trans}
The transferability of an attack measures the capability of adversarial examples crafted from one model misleading another one.
It is important to compare the transferability of different attacks, which shows the high potential to realize effective black-box attacks.
For a fair comparison, we conduct the transferability experiments of all compared methods under the same image quality.
%
{Specifically, we select the maximum perturbation as $8$ for six attacks (\ie, FGSM/MIFGSM/DIM/TIFGSM/TIMIFGSM/TIDIM), the weight $c$ as $1.0$ for C\&W, the maximum perturbation pixels number $k$ as $10000$ for SA since they get similar image quality with our method under this setup.}
We use the same setup for both DEV and CVPR2021-AIC-VI datasets.

We craft adversarial examples from Inc-v3, Inc-v4, and IncRes-v2 and feed them to fool all three models and ResNet50 \cite{he2016deep}, respectively, then we obtain twelve attack results for each attack method.
As shown in Table~\ref{Tab-noisecompare} and \ref{Tab-devtianchi} for DEV and CVPR2021-AIC-VI datasets, \textit{Pasadena} achieves the highest attack and transferability among all compared baselines, followed by DIM and C\&W.
More specifically, when considering the adversarial examples crafted from Inc-v3 on DEV dataset, C\&W has the second-highest attack success rate of $99.5\%$ while our method gets the best result with $100.0\%$. 
Moreover, in terms of the transferability on DEV dataset, \textit{Pasadena} gets $78.2\%$, $75.8\%$ and $72.7\%$ success rates on ResNet50, Inc-v4 and IncRes-v2, respectively, with the adversarial examples crafted from Inc-v3, which are significantly higher than the results of the best baseline, \ie, DIM, with success rates of $52.3\%$, $48.3\%$ and $49.2\%$, respectively. We see greater advantages on the CVPR2021-AIC-VI dataset.
The advantages of our method mainly stem from the proposed kernel prediction-based attack and our perceptual region localization.
In particular, the kernel prediction-based attack distorts the input image by linearly combing local neighboring pixels, which is fundamentally different from the additive-perturbation-based attacks modifying each pixel independently. The local combination way makes more pixels contribute to the distortions of high-level features, thus can achieve higher transferability.
%
Moreover, we use the perceptual region localization to find the vulnerable regions that let the attack be more effective while keeping high image quality.
%
%
In general, our \textit{Pasadena} has the best attack capability and transferability under the same image quality.

\subsection{Comparison with Denoise~\&~Attack Baselines}
\label{subsec:comp_baselines_wadenoi}

\begin{figure*}
    \centering
    \subfigure { 
    \includegraphics[width=0.31\linewidth]{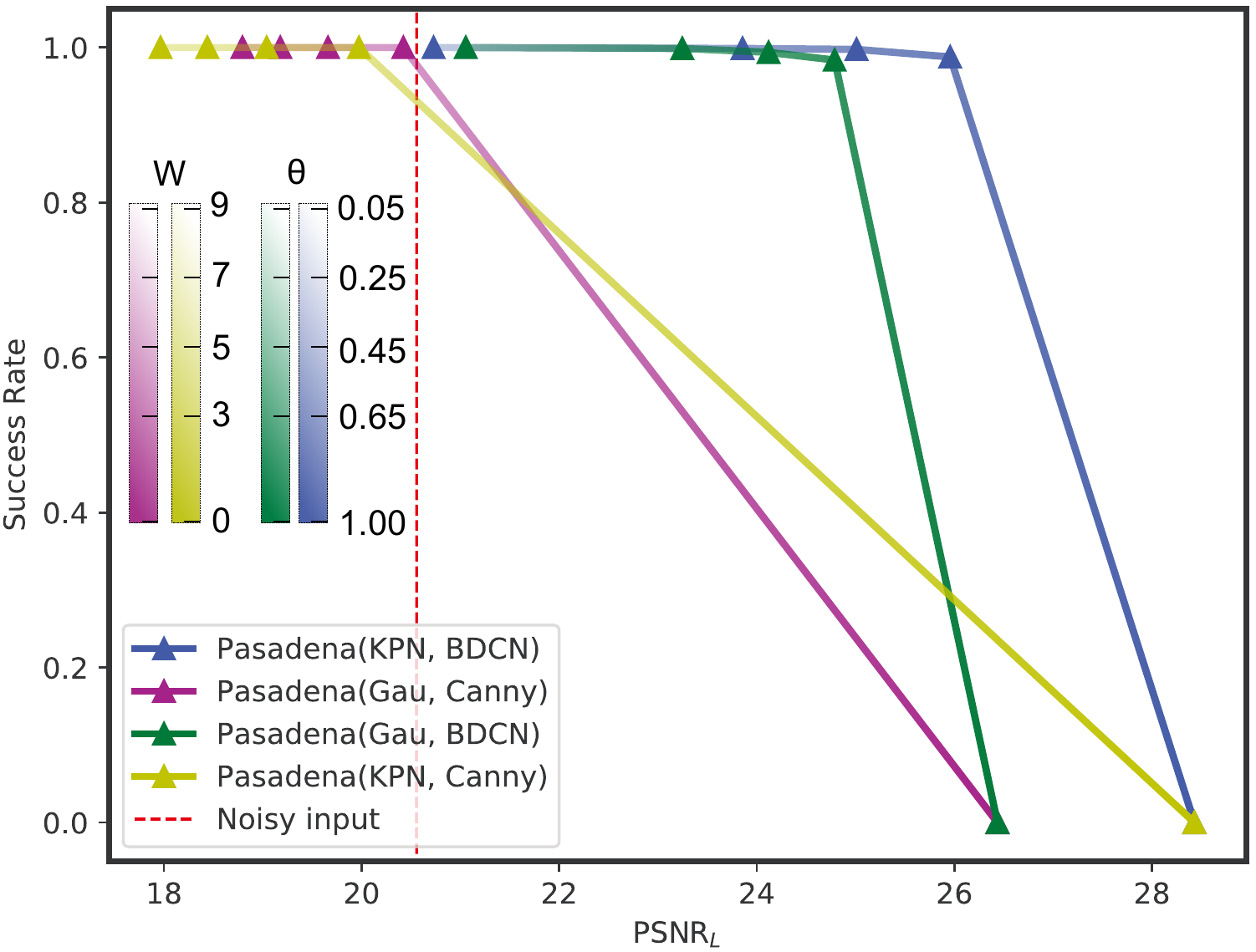}
    } 
    \subfigure { 
    \includegraphics[width=0.31\linewidth]{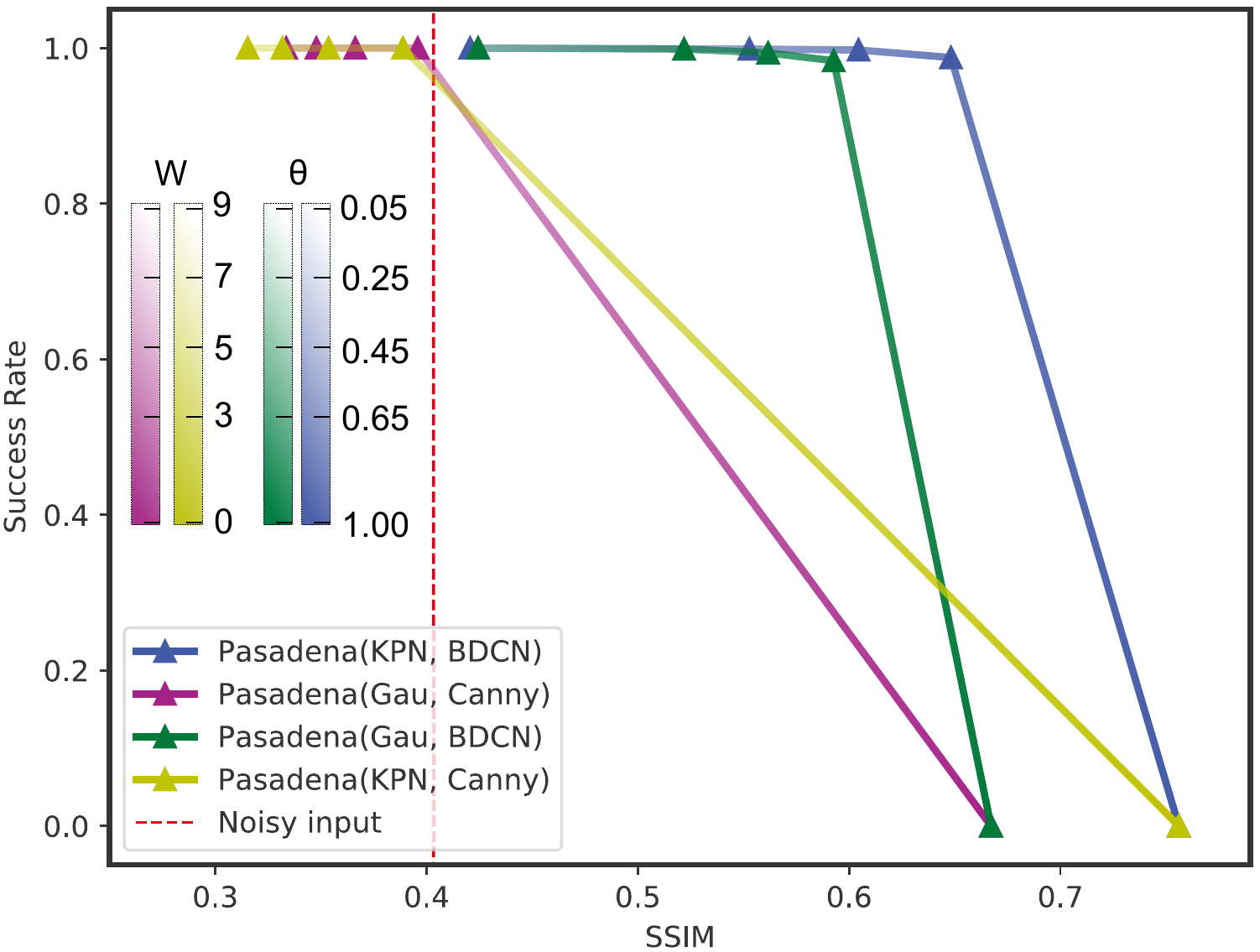}
    } 
    \subfigure { 
    \includegraphics[width=0.31\linewidth]{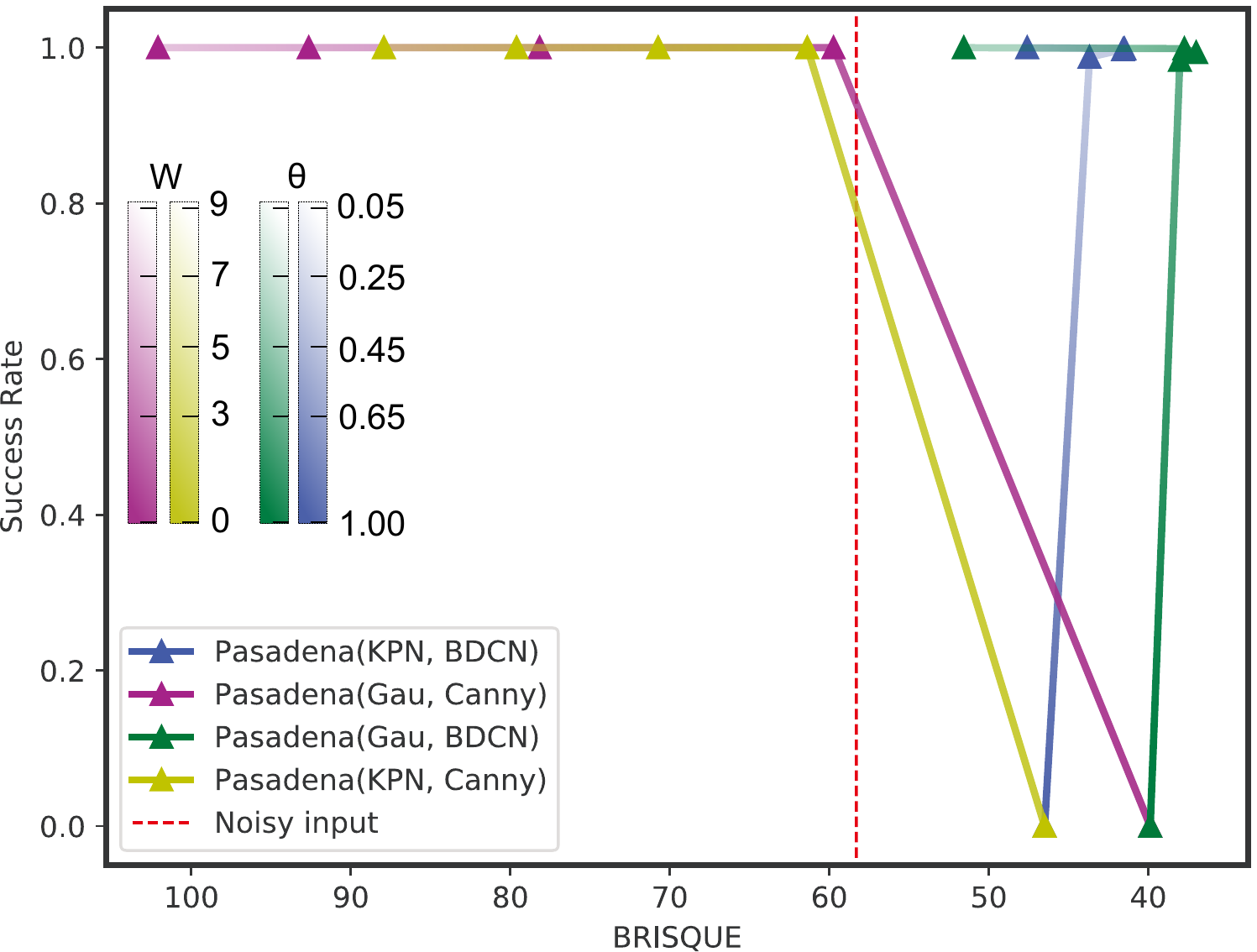}
    } 
    \caption{Whitebox attack success rate along with $\mathrm{PSNR}_{L}$, SSIM, and BRISQUE for our framework with a different choice of denoising and edge detector methods. \textbf{KPN} refers to the denoising method proposed in \cite{Mildenhall2018CVPR}. \textbf{Gau} means Gaussian kernel is selected in our denoising part. \textbf{Canny} and \textbf{BDCN} are edge detecting methods of \cite{canny1986computational} and \cite{He2019CVPR}, respectively. The dots with zero success rate represent the performance of denoising results using \textbf{KPN} (blue and yellow) and Gaussian kernel (green and purple). The red dash line represents the quality of noisy input. The color gradient refers to the change of $W$ and $\theta$ mentioned in Sec.~\ref{subsec:Ablation_wa}.
    }
\label{fig:exp_Ablation}
\end{figure*}   

\subsubsection{Quantitative Analysis}
%
In this section, we consider eight more challenging baselines that first process a noisy image with a denoising method and then attack the denoised image with the eight attack baselines. 
Here, we use the denoising method called kernel-prediction-based denoising \cite{Mildenhall2018CVPR} in Pasadena, for a fair comparison.
All settings are similar with Sec.~\ref{subsec:comp_baselines_wa} and the results are shown in Fig.~\ref{fig:exp_denoise}.
We have the following observations:
{
\textit{First}, in contrast to the attack baselines in Sec.~\ref{subsubsec:comp_baselines_wa_quatitative}, the denoise~\&~attack methods can enhance the image quality of the noisy inputs by pre-processing them with the denoising method.
However, \textit{Pasadena} still achieves much better image quality than six new baselines, \ie, Denoise+FGSM/MIFGSM/DIM/TIFGSM/TIMIFGSM/TIDIM, under similar success rates. 
Despite two challenging baselines, \ie, C\&W and SA, outperform our method, they show weakness on transferability which will be discussed in the upcoming section (Sec. \ref{subsec:comp_denoise_baselines_trans}).
Specifically, as shown in Fig.~\ref{fig:exp_denoise}, considering the success rate around $96\%$, we see that three variants of \textit{Pasadena} with $\lambda=0.9, 0.5, 0.3$ obtain much higher \PSNRL{} and \SSIM{} while lower \BRIS{} than six of all compared denoise~\&~attack methods. But, denoise~\& C\&W and denoise~\& SA beat other methods as they consider the image quality in the attack process.
}
%
%
%

{
\textit{Second}, in terms of the success rate under similar image quality, \textit{Pasadena} keeps the superiority over most of the denoise~\&~attack methods. For example, when the \SSIM{} values are around $0.6$, Pasadena achieves $100\%$ attack success rate while the six baselines, \ie, FGSM/MIFGSM/DIM/TIFGSM/TIMIFGSM/TIDIM, get success rates ranging from $40\%$ to $98\%$.
}

\subsubsection{Qualitative Analysis}
Similar to Fig.~\ref{fig:via_baselines} in Sec.~\ref{subsubsec:comp_baselines_wa_qualitative}, Fig.~\ref{fig:via_baselines_denoise} visualizes six adversarial attack images of Pasadena and eight denoise~\&~attack methods. 
In contrast to the attack baseline methods that further corrupt the noisy inputs, the denoise~\&~attack baselines can improve the image quality and get much clearer visualization results than the noisy inputs but still embed adversarial noise pattern, \eg, the results of Denoise+TIFGSM/TIMIFGSM/TIDIM/C\&W/SA.
Nevertheless, our method achieves much better denoise effects.
For example, the `Panda' case in the first row shows that our method maintains the denoising effect for non-edge regions (\eg, the face and body of the panda), and only implement attacks on the edge regions (\eg, the boundary across the black and white regions), leading to successful attack (\eg, classifier predicts the incorrect category `Bobtail') with better perceptual visualization (\ie, higher \PSNRL{}, \SSIM{}, and lower \BRIS{}).
In contrast, denoise~\&~attack methods perturb the entire denoised image and result in re-corrupted images with obvious noise textures. Specifically, the Denoise+TIFGSM/TIMIFGSM/TIDIM/C\&W methods produce salient noise textures on all images even though they have similar \PSNRL{} and \SSIM{} with our method.
{Denoise+SA mainly perturbs the panda's face while resulting in obvious perturbations, even though it has the best \PSNRL{} and \SSIM{} values.} Moreover, some baselines fail to mislead DNNs. 
For example, Denoise+FGSM/TIFGSM does not change the predicted category of the input `basketball player'.

\subsubsection{Comparison on Transferability}
\label{subsec:comp_denoise_baselines_trans}

To further compare the transferability of Pasadena with denoise~\&~attack methods, we conduct an experiment with the same set of Sec.~\ref{subsec:comp_baselines_trans}. In contrast, we select the weight $c$ as $10.0$ for the C\&W method and the maximum perturbation pixels number $k$ as $30000$ for the SA method since they share a similar image quality, as shown in Fig.~\ref{fig:exp_denoise}. 
We show the attack success rates as well as the transferability results across four normally trained models in Table~\ref{Tab-denoisecompare}.

Generally, our \textit{Pasadena} still maintains the highest attack capability and transferability for all subject models, \ie, Inc-v3, Inc-v4, and IncRes-v2, followed by Denoise+C\&W.
%
To be more specific, in terms of the adversarial attacking results crafted from Inc-v3, Denoise+C\&W keeps the same highest attack success rate of $100\%$. Nevertheless, its success rates of using adversarial examples to attack ResNet50, Inc-v4 and IncRes-v2, \ie, $48.4\%$, $46.5\%$ and $48.5\%$, are significantly lower than those of our method, \ie, $78.2\%$, $75.8\%$ and $72.7\%$. 

There is an interesting observation: comparing the results in Table~\ref{Tab-noisecompare} and \ref{Tab-denoisecompare}, we find that part of the attacks, \ie, MIFGSM/DIM/TIMIFGSM/TIDIM, have better attack capability while worse transferability on denoised inputs than on noisy inputs. To be specific, taking the adversarial images crafted from Inc-v3 for example, DIM has the attack success rates of $94.3\%$ for noisy input and $97.3\%$ for denoised input. Yet, in terms of the transferability, the attack success rate drops from $49.2\%$ (noisy input) to $36.7\%$ (denoised input) for attacking IncRes-v2. 
This observation shows that, for part of the attack baselines, denoising procedure benefits attack capability while harming the transferability in some cases.
In contrast to the baselines' sensitivity to the denoising method, our \textit{Pasadena} achieves the strongest attack capability and transferability. 
%
We find similar adversarial results crafted for the other two models, which draw the same conclusion as in Sec.~\ref{subsec:comp_baselines_trans}, \ie, our adversarial attacking results have the highest transferability over all baselines.

\begin{table*}[t]
\centering
\small
\caption{ Success rates of four variants of our method with different denoising methods and edge detectors against three normally trained models: Inc-v3, Inc-v4, and IncRes-v2. For each compared group (\ie, the three columns for Inc-v4, IncRes-v2, and Inc-v3), whitebox attack results are shown in the first one. The rest two columns display the blackbox attack results. 
We highlight the best two results with red and green fonts, respectively.
}
\label{Tab-ablation}
\begin{tabular}{l|ccc|ccc|ccc}
\toprule
\rowcolor{tabgray} \multicolumn{1}{c}{Crafted from} & \multicolumn{3}{c}{Inc-v3} & \multicolumn{3}{c}{Inc-v4} & \multicolumn{3}{c}{IncRes-v2} \\
\cmidrule(r){1-1} \cmidrule(r){2-4} \cmidrule(r){5-7} \cmidrule(r){8-10}
\rowcolor{tabgray} Attacked model
&  Inc-v3   &   IncRes-v2 &  Inc-v4
&  Inc-v4   &   IncRes-v2 &  Inc-v3
&  IncRes-v2      &  Inc-v4 &   Inc-v3\\
\midrule
Pasadena~(Gau, Canny)  
&99.9           &63.8  &68.0  
&100.0  &65.3  &71.1 
&100.0  &74.6  &74.5       \\
%
%
Pasadena~(KPN, Canny) 
&100.0  &68.6           &\second{73.3}   
&100.0  &70.8           &72.4 
&99.8           &\second{80.1}  &77.5        \\
Pasadena~(Gau, BDCN) 
&100.0   &\second{70.1}     &72.9    
&100.0   &\second{71.6}     &\second{72.5} 
&100.0   &79.5              &\second{78.3}        \\
%
%
Pasadena~(KPN, BDCN)  
&100.0  &\first{72.7} &\first{75.8}   
&100.0  &\first{74.1} &\first{72.7} 
&100.0  &\first{81.3} &\first{78.4}         \\
\bottomrule
\end{tabular}
\end{table*}

\subsection{Effects of Denoisers and Edge Detectors}
\label{subsec:Ablation_wa}
In this section, we present the results of our framework with different setups of denoising and edge detector methods.
To be specific, we choose Gaussian filter (Gau), and, KPN \cite{Mildenhall2018CVPR}, for the denoising module, and Canny operator \cite{canny1986computational}, and BDCN \cite{He2019CVPR} for the edge detecting module.
For clear representation and discussion, we denote four variants of Pasadena with different denoising methods and edge detectors as Pasadena~(Gau, Canny), Pasadena~(KPN, Canny), Pasadena~(Gau, BDCN), and Pasadena~(KPN, BDCN).
We compare the quantitative results in Table~\ref{Tab-ablation} and Fig.~\ref{fig:exp_Ablation}. 
In general, we find that our method with the very basic image denoising method, \eg, Gaussian filter, and edge detector, \eg, Canny, can still obtain a significantly high attack success rate and transferability, while the advanced denoiser and detector contribute to better performance.

\subsubsection{Quantitative Analysis}

\begin{figure}
    \centering
    \includegraphics[width=1.0\linewidth]{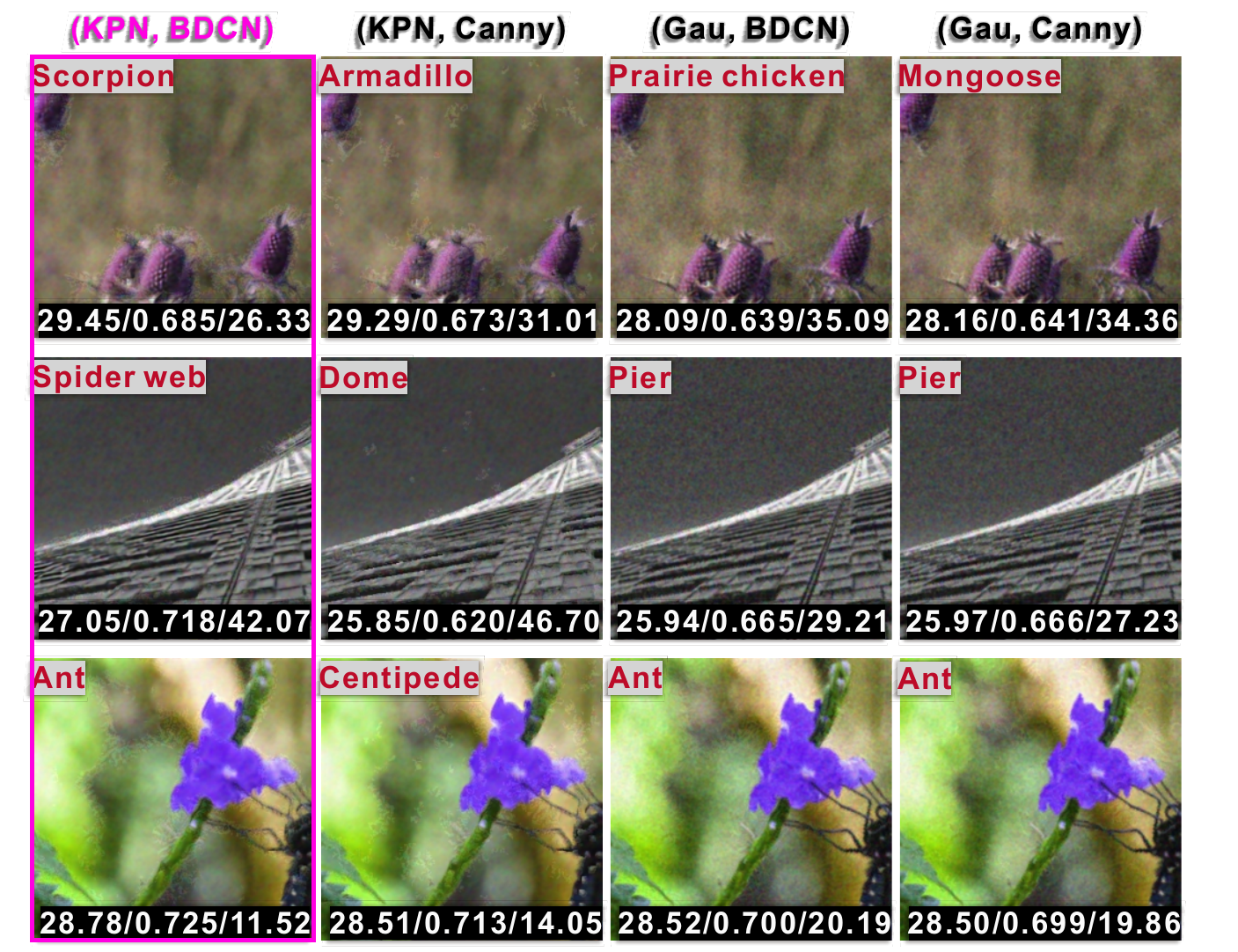}
    \caption{Visualization results of adversarial examples crafted for Inception-v4, using our method Pasadena with different denoising methods and edge detectors. The denoising methods are KPN \cite{Mildenhall2018CVPR} and Gaussian kernel (kernel size=5, $\sigma=1$), while edge detectors are BDCN and Canny. The combination is listed on top of each group. For each image, the prediction of Inception-v4 is displayed on the top-left. Note that, all attacks succeed. Three numbers at the bottom refer to $\mathrm{PSNR}_{L}$, SSIM, and BRISQUE.}
\label{fig:via_ablation}
\end{figure}

Fig.~\ref{fig:exp_Ablation} illustrates the performance of our four variants and we see that:
more powerful edge detectors and image denoiser help realize higher image quality under the same hyper-parameters. Specifically, as shown in Fig.~\ref{fig:exp_Ablation}, with the same denoising method (\ie, KPN, Gau) and the same hyper-parameters, Pasadena~(KPN/Gau, BDCN) can realize a significantly higher success rate with much better image quality than Pasadena~(KPN/Gau, Canny). For example, under the same $\theta=[0.05, 0.65]$, Pasadena~(KPN, Canny) has the \SSIM{} ranging from $0.3$ to $0.4$ while Pasadena~(KPN, BDCN) achieves much higher values from $0.4$ to $0.67$. 
Similarly, under the same edge detector (\ie, Canny or BDCN) and hyper-parameters, Pasadena~(KPN, BDCN/Canny) can realize much better image quality than Pasadena~(Gau, BDCN/Canny). 

\subsubsection{Qualitative Analysis}
We list three groups of adversarial examples using a different combination of denoising methods and edge detectors in Fig.~\ref{fig:via_ablation} for visual comparison. 
It is easy to see that the combination of KPN and BDCN successfully attacks the inputs as well as keeping the largest \PSNRL{} and \SSIM{} values among the four groups. This combination also has an acceptable performance when considering the \BRIS{} metric. 
%
{To be specific, KPN-based methods achieve clearer background in the case `Scorpion' and `Spider web' than the Gaussian kernel-based denoising, indicating the advantages of KPN for adversarial noise attack. 
In addition, comparing the attack area of the second input, Pasadena~(KPN, Canny) pollutes the sky area, while Pasadena~(Gau, BDCN) only locates the attack pattern in the tile roof part. That means BDCN can precisely locate vulnerable areas while keeping the denoising effect for the non-edge part.}

%

\subsubsection{Comparison on transferability}
\label{subsec:Ablation_trans}
To further study the influence of the denoisers and edge detectors on the transferability, we use a similar hyper-parameters selection strategy as the former experiment in Sec~\ref{subsec:comp_baselines_trans} for our four variants (See the most left column of Table~\ref{Tab-ablation}).
For the parameters of Canny and BDCN, we choose $W=3$ for Pasadena~(KPN/Gau, Canny) and $\theta=0.05$ for Pasadena~(KPN/Gau, BDCN) since they share similar SSIM values near $0.4$ in Fig~\ref{fig:exp_Ablation}.
Table~\ref{Tab-ablation} shows the results. In general, all combinations achieve almost $100.0\%$ success rates under the white-box configuration.
In terms of transferability, the variant with KPN and BDCN has the best performance. 
The adversarial examples crafted from Inc-v4 by Pasadena~(KPN, BDCN) have success rates of $75.8\%$ and $72.7\%$ when we use them to attack Inc-v4 and IncRes-v2, respectively, followed by $72.5\%$ and $71.6\%$ of Pasadena~(Gau, BDCN). This conclusion further proves that better denoising and edge detecting results enhance our final attacking performance.
Intuitively, denoising via KPN benefits high image qualities while BDCN is able to find more vulnerable boundary areas. In contrast, Canny edge detection tends to localize a lot of less effective targeted areas. Attacking these areas sacrifices the image quality while failing to increase the attack success rate.

\begin{table*}[t]
\centering
\small
\caption{Adversarial comparison results on DEV dataset with additive Gaussian noise ($\sigma$=0.1). It contains the success rates (\%) of eight baseline attack methods and our Pasadena against the adversarially trained ResNet50 (\ie, ResNet50$_\text{at}$) \cite{wong2020fast}.
We highlight the best two results with red and green fonts, respectively.}
\label{Tab-defensemodel}
\begin{tabular}{l|c|c|c|c|c|c|c|c}
\toprule
\rowcolor{tabgray}{Crafted from} & ResNet50$_\text{at}$ & {Inc-v3} & {Inc-v4} & {IncRes-v2} & ResNet50 & {Inc-v3} & {Inc-v4} & {IncRes-v2} \\
\rowcolor{tabgray}{Attacked model} & \multicolumn{4}{c|}{ResNet50$_\text{at}$} & \multicolumn{4}{c}{ResNet50}\\
\midrule
FGSM              &82.5  &36.5    &37.4   &37.3        
                  &99.0  &44.0    &47.4   &47.9   
\\
TIFGSM            &76.3  &\second{39.7}    &39.8   &\second{40.6}       
                  &95.2  &45.0    &46.0   &48.2\\
MIFGSM            &\second{90.6}  &35.2    &35.9   &35.7       
                  &\first{100.0}  &34.2    &39.8   &46.9       \\
TIMIFGSM          &82.0  &38.0     &38.8    &39.6       
                  &\first{100.0}  &41.4    &45.3   &52.9        \\
DIM               &86.2  &36.2    &36.8   &36.9       
                  &\first{100.0}  &52.3    &\second{56.6}   &\second{61.9}        \\
TIDIM             &76.2  &38.9     &\second{40.1}    &40.5       
                  &99.1  &\second{54.6}    &55.7   &61.8       \\
C\&W ($L_2$ norm) &86.8  &28.8     &30.4    &31.3     
                  &\first{100.0}  &26.2    &27.5   &29.2                   \\
SA                &64.6  &34.4     &33.9    &34.0      
                  &95.8  &25.1    &25.7   &26.1               \\
Pasadena (ours)   &\first{98.5}  &\first{51.3}     &\first{53.2}    &\first{55.4}     
                  &\second{99.9}  &\first{78.2}    &\first{78.0}   &\first{81.0}                 \\
\bottomrule
\end{tabular}
\end{table*}
%

\subsection{Attack Results on Defense Models}
\label{subsec:attack_on_defense_models}

Besides the attack results against three standard-trained models, we further evaluate our adversarial attacks as well as baselines on the adversarial-trained ResNet50 (ResNet50$_\text{at}$) model introduced in \cite{wong2020fast}. We also add the attack results on the standard-trained ResNet50 as a comparison.
Note that, ResNet50$_\text{at}$ is trained via the adversarial training with the FGSM to generate adversarial examples on the fly during the training process.
Specifically, we use attacking methods to generate adversarial examples crafted from ResNet50$_\text{at}$, Inc-v3, Inc-v4, and IncRes-v2 and attack the ResNet50$_\text{at}$. 
We can conduct the same experiments on ResNet50.
The attack success rates are reported in Table~\ref{Tab-defensemodel}. In general, the attack success rate on ResNet50$_\text{at}$ is lower than the results on ResNet50 under white-box attacks, which demonstrates the adversarial training does enhance the robustness of ResNet50.
Compared with baseline attack methods, \textit{Pasadena} achieves the highest attack success rate under both white-box and transfer-based attacks.
For example, considering the adversaries crafted from ResNet50$_\text{at}$, \textit{Pasadena} has an attack success rate of $98.5\%$, much higher than two SOTA baselines, \ie, DIM and TIDIM, which are $86.2\%$ and $76.2\%$. Moreover, C\&W and SA are also heavily affected by the defense model as they sacrifice attack capability for higher image quality.
%
This experiment demonstrates that recent defense models have defencing effect against attacks. However, rather than only considering noisy adversarial attacks, its improvement needs to take other types of perturbation like our adversarial denoise attack into account.

%

\subsection{Influence of noise level}
\begin{figure}
    \centering
    \includegraphics[width=0.9\linewidth]{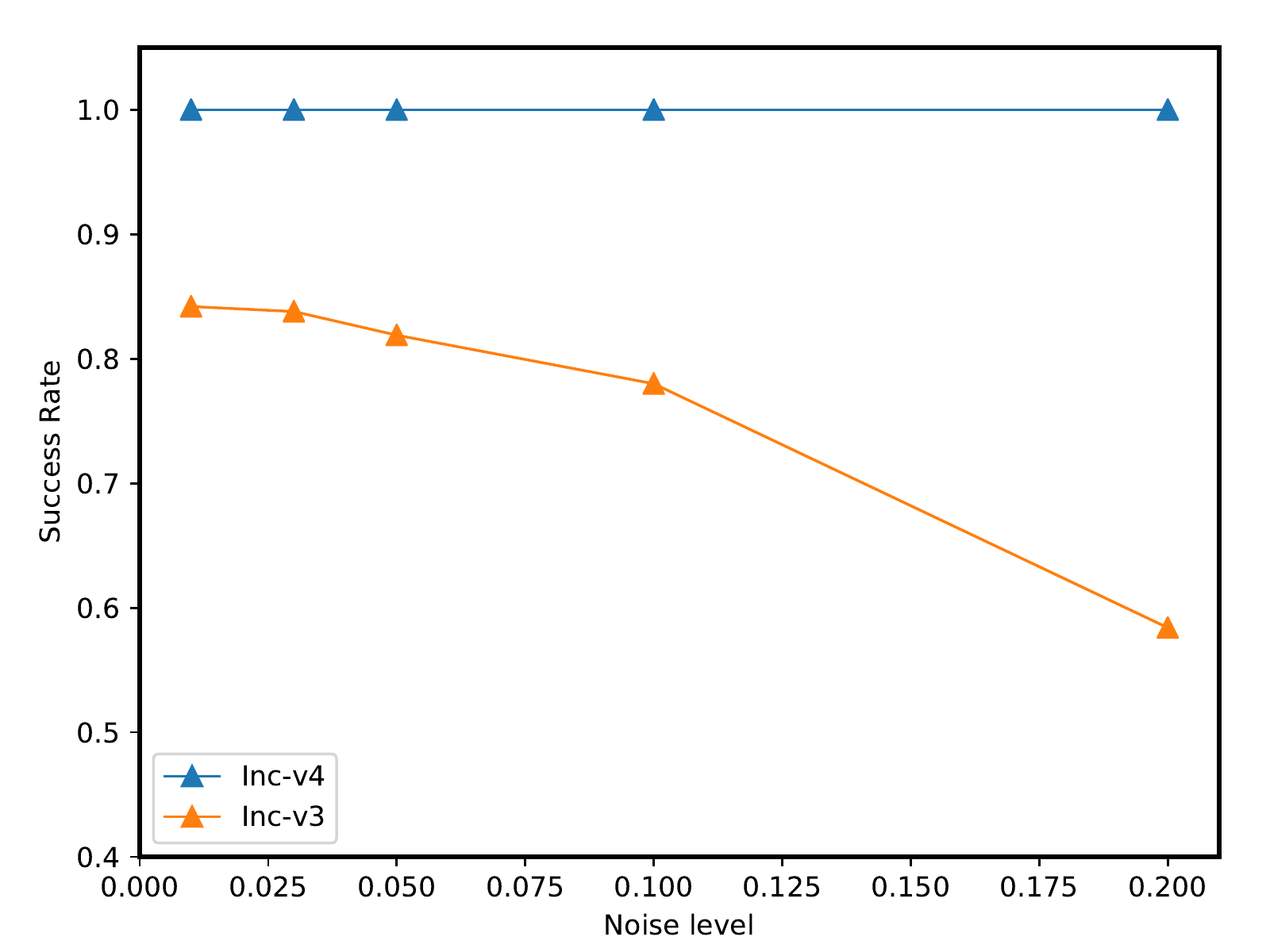}
   
    \caption{We attack Inc-v4 and Inc-v3 models with adversarial examples crafted from Inc-v4 model.  
    }
\label{fig:multi_noise}
\end{figure}

To evaluate the performance of our method in different noise conditions, we conduct adversarial attacks against the Inc-v4 model on NeurIPS’17 adversarial competition dataset with different additive Gaussian noise ($\sigma=0.01, 0.03, 0.05, 0.1, 0.2$). Moreover, we also use these adversarial examples to fool the Inc-v3 model, testing the influence of noise on our transferability. 
The results are shown in Fig.~\ref{fig:multi_noise}.
With the increase of noise level, the success rate when attacking Inc-v4 maintains $100\%$. This means that our attack keeps high attack capability even in heavy noise conditions. However, our transferability, \ie,  the success rate when attacking Inc-v3, drops from $84\%$ to $58\%$ in heavy noise. It indicates that the random noise may have a negative effect against adversarial attacks.

\subsection{Attack results on other noise types}
\label{subsec:other_degradation_dataset}
In this subsection, we further conduct the adversarial denoise attack on other three noise types, \ie, shot noise, impulse noise, and speckle noise, to validate the generalization of our method. 
Shot noise is also known as Poisson noise caused by the discrete nature of light while impulse noise is a color analog of salt-and-pepper noise \cite{hendrycks2019robustness}. 
Speckle noise is multiplicative and usually exists in medical images.
We use our method to attack two powerful deep models, \ie, ResNet-101 \cite{he2016deep} and EfficientNet \cite{tan2019efficientnet}, on the three noise subsets of Tiny-ImageNet-C~\cite{hendrycks2019robustness} where each noise type contains five severity levels. The two DNNs are pre-trained on Tiny-ImageNet and we also evaluate the transferability by using the adversarial examples from one DNN to attack another one.
We present the attacking results in Tabel~\ref{Tab-othernoise} and have the following observations: \textit{First}, our method still works for the three noise, that is, we can achieve a significantly high attack success rate while improving the image quality (\ie, SSIM vs. Original SSIM in Table~\ref{Tab-othernoise}) across different severity levels. For example, when attacking ResNet-101 at severity level two, we get 74.8\% and 74.0\% attack success rates on shot noise and impulse noise, respectively, while the SSIMs increase from 0.735 and 0.754 to 0.790 and 0.785, respectively. In addition, the image quality enhancements become more significant as the severity level increases.
For the speckle noise at the first severity level, the adversarial examples have similar quality with the original noisy images. At other severity levels, the image quality is improved significantly after attacking.
\textit{Second}, for the transferability, our method achieves similar transferability across different severity levels with around 30\% success rates, which are high scores under transfer-based attacks. 

\begin{table*}[t]
\centering
\small
\caption{Adversarial comparison results on Tiny-imagenet-C dataset with shot noise, impulse noise and speckle noise. It contains the success rates (\%) of black \& whitebox adversarial attacks among two normally trained models, Resnet-101 and EfficientNet. The attacks are conducted on 5 severity levels. The third column lists the original SSIM values of the perturbed dataset before attack.}
\label{Tab-othernoise}
{
\begin{tabular}{l|c|c|ccc|ccc}
\toprule
 \multirow{2}{*}{Noise types}  & \multirow{2}{*}{\makecell{Severity\\Level}} & \multirow{2}{*}{\makecell{Original\\SSIM}}  & \multicolumn{3}{c|}{ResNet-101} & \multicolumn{3}{c}{EfficientNet}\\
%
%
& & &  ResNet-101   &   EfficientNet  &   SSIM

&  ResNet-101   &   EfficientNet  &   SSIM\\
\midrule
\multirow{5}{*}{Shot noise} 
           & 1 & 0.828 & 82.3  & 25.6  & 0.844  & 23.7   &84.8  & 0.848  \\
           & 2 & 0.735 & 74.8  & 25.5  & 0.790  & 24.1   &80.9  & 0.794    \\
           & 3 & 0.639 & 62.0  & 24.0  & 0.714  & 25.1   &70.1  & 0.719   \\
           & 4 & 0.563 & 54.0  & 24.4  & 0.644  & 27.7   &60.1  & 0.649    \\
           & 5 & 0.462 & 47.6  & 23.7  & 0.540  & 31.0   &52.6  & 0.543    \\
\midrule
\multirow{5}{*}{Impulse noise} 
           & 1 & 0.833 & 81.8  & 29.0  & 0.833 & 27.5   & 86.0 & 0.837  \\
           & 2 & 0.754 & 74.0  & 28.8  & 0.785 & 31.6   & 81.6 & 0.787    \\
           & 3 & 0.593 & 59.9  & 28.1  & 0.660 & 40.7   & 68.2 & 0.661   \\
           & 4 & 0.504 & 53.0  & 26.2  & 0.579 & 44.9   & 59.3 & 0.581    \\
           & 5 & 0.389 & 44.4  & 26.6  & 0.464 & 42.5   & 49.6 & 0.465    \\
\midrule
\multirow{5}{*}{Speckle noise} 
           & 1 & 0.869 & 80.3  & 20.9  & 0.865 & 21.5   & 84.1 & 0.868  \\
           & 2 & 0.640 & 70.2  & 24.6  & 0.709 & 34.2   & 74.4 & 0.710    \\
           & 3 & 0.534 & 57.6  & 24.4  & 0.631 & 36.2   & 62.7 & 0.628   \\
           & 4 & 0.471 & 54.7  & 25.4  & 0.584 & 38.8   & 57.8 & 0.576    \\
           & 5 & 0.429 & 54.6  & 26.0  & 0.546 & 39.0   & 55.4 & 0.539    \\
\bottomrule
\end{tabular}
}
\end{table*}
\section{Conclusions and Discussion}\label{sec:conc}

In this work, we have investigated a new task named the \textit{adversarial denoise attack} that stealthily embeds attacks inside the image denoising module. Thus, it can simultaneously denoise the input images while fooling the state-of-the-art deep models. We have formulated this new task as a kernel prediction problem for image filtering and proposed the \textit{adversarial-denoising kernel prediction} that can produce adversarial-noiseless kernels for effective denoising and adversarial attacking simultaneously. Furthermore, we have implemented an adaptive \textit{perceptual region localization} to identify the semantic-related vulnerability regions with which the attack can be more effective while not doing too much harm to the denoising. As a result, our proposed method is termed as \textit{Pasadena} (Perceptually Aware and Stealthy Adversarial DENoise Attack). We have validated our method on the NeurIPS'17 adversarial competition dataset, CVPR2021-AIC-VI:unrestricted adversarial attacks on ImageNet, and Tiny-ImageNet-C dataset, and demonstrated that our method not only realizes denoising but also has advantages of high success rate and transferability over the state-of-the-art attacks. 

In the future, we could extend our attack by considering other kinds of image degradation, \eg, blur \cite{iccv21_advmot}, rain \cite{zhai2020s}, haze \cite{gao2021advhaze}, \etc, and conduct adversarial deblurring/deraining/dehazing attacks. Moreover, we could construct a unified adversarial image-restoration attack that can address diverse degradation in a single framework.

\bibliographystyle{IEEEtran}
\bibliography{ref}

\begin{thebibliography}{10}
\providecommand{\url}[1]{#1}
\csname url@samestyle\endcsname
\providecommand{\newblock}{\relax}
\providecommand{\bibinfo}[2]{#2}
\providecommand{\BIBentrySTDinterwordspacing}{\spaceskip=0pt\relax}
\providecommand{\BIBentryALTinterwordstretchfactor}{4}
\providecommand{\BIBentryALTinterwordspacing}{\spaceskip=\fontdimen2\font plus
\BIBentryALTinterwordstretchfactor\fontdimen3\font minus
  \fontdimen4\font\relax}
\providecommand{\BIBforeignlanguage}[2]{{%
\expandafter\ifx\csname l@#1\endcsname\relax
\typeout{** WARNING: IEEEtran.bst: No hyphenation pattern has been}%
\typeout{** loaded for the language `#1'. Using the pattern for}%
\typeout{** the default language instead.}%
\else
\language=\csname l@#1\endcsname
\fi
#2}}
\providecommand{\BIBdecl}{\relax}
\BIBdecl

\bibitem{tian2020deep}
C.~Tian, L.~Fei, W.~Zheng, Y.~Xu, W.~Zuo, and C.-W. Lin, ``Deep learning on
  image denoising: An overview,'' \emph{Neural Networks}, 2020.

\bibitem{cho2017geodesic}
S.~I. Cho and S.-J. Kang, ``Geodesic path-based diffusion acceleration for
  image denoising,'' \emph{IEEE Transactions on Multimedia}, vol.~20, no.~7,
  pp. 1738--1750, 2017.

\bibitem{malladi2020image}
S.~R.~S. Malladi, S.~Ram, and J.~J. Rodriguez, ``Image denoising using
  superpixel-based pca,'' \emph{IEEE Transactions on Multimedia}, 2020.

\bibitem{ghosh2019fast}
S.~Ghosh and K.~N. Chaudhury, ``Fast bright-pass bilateral filtering for
  low-light enhancement,'' in \emph{2019 IEEE International Conference on Image
  Processing (ICIP)}.\hskip 1em plus 0.5em minus 0.4em\relax IEEE, 2019, pp.
  205--209.

\bibitem{he2016deep}
K.~He, X.~Zhang, S.~Ren, and J.~Sun, ``Deep residual learning for image
  recognition,'' in \emph{Proceedings of the IEEE Conference on Computer Vision
  and Rattern Recognition}, 2016, pp. 770--778.

\bibitem{dong2017cunet}
L.~Dong, L.~He, M.~Mao, G.~Kong, X.~Wu, Q.~Zhang, X.~Cao, and E.~Izquierdo,
  ``Cunet: A compact unsupervised network for image classification,''
  \emph{IEEE Transactions on Multimedia}, vol.~20, no.~8, pp. 2012--2021, 2017.

\bibitem{zhang2019unsupervised}
C.~Zhang, J.~Cheng, and Q.~Tian, ``Unsupervised and semi-supervised image
  classification with weak semantic consistency,'' \emph{IEEE Transactions on
  Multimedia}, vol.~21, no.~10, pp. 2482--2491, 2019.

\bibitem{guo2019distributed}
Y.~Guo, B.~Zou, J.~Ren, Q.~Liu, D.~Zhang, and Y.~Zhang, ``Distributed and
  efficient object detection via interactions among devices, edge, and cloud,''
  \emph{IEEE Transactions on Multimedia}, vol.~21, no.~11, pp. 2903--2915,
  2019.

\bibitem{cong2018hscs}
R.~Cong, J.~Lei, H.~Fu, Q.~Huang, X.~Cao, and N.~Ling, ``Hscs: Hierarchical
  sparsity based co-saliency detection for rgbd images,'' \emph{IEEE
  Transactions on Multimedia}, vol.~21, no.~7, pp. 1660--1671, 2018.

\bibitem{Liu2018IJCAI}
D.~Liu, B.~Wen, X.~Liu, Z.~Wang, and T.~Huang, ``When image denoising meets
  high-level vision tasks: A deep learning approach,'' in \emph{Proceedings of
  the 27th International Joint Conference on Artificial Intelligence, IJCAI
  2018}, 2018, pp. 842--848.

\bibitem{Szegedy17Incv4}
C.~Szegedy, S.~Ioffe, V.~Vanhoucke, and A.~A. Alemi, ``Inception-v4,
  inception-resnet and the impact of residual connections on learning,'' in
  \emph{AAAI}, ser. AAAI’17, 2017, p. 4278–4284.

\bibitem{krasin2017openimages}
I.~Krasin, T.~Duerig, N.~Alldrin, V.~Ferrari, S.~Abu-El-Haija, A.~Kuznetsova,
  H.~Rom, J.~Uijlings, S.~Popov, A.~Veit \emph{et~al.}, ``Openimages: A public
  dataset for large-scale multi-label and multi-class image classification,''
  \emph{Dataset available from https://github. com/openimages}, vol.~2, no.~3,
  pp. 2--3, 2017.

\bibitem{Mildenhall2018CVPR}
B.~{Mildenhall}, J.~T. {Barron}, J.~{Chen}, D.~{Sharlet}, R.~{Ng}, and
  R.~{Carroll}, ``Burst denoising with kernel prediction networks,'' in
  \emph{Proceedings of the IEEE Conference on Computer Vision and Pattern
  Recognition}, 2018, pp. 2502--2510.

\bibitem{dong2019evading}
Y.~Dong, T.~Pang, H.~Su, and J.~Zhu, ``Evading defenses to transferable
  adversarial examples by translation-invariant attacks,'' in \emph{Proceedings
  of the IEEE Conference on Computer Vision and Pattern Recognition}, 2019, pp.
  4312--4321.

\bibitem{DFDC2020}
B.~Dolhansky, J.~Bitton, B.~Pflaum, J.~Lu, R.~Howes, M.~Wang, and C.~C. Ferrer,
  ``The deepfake detection challenge dataset,'' 2020.

\bibitem{ijcai20_fakespotter}
R.~Wang, F.~Juefei-Xu, L.~Ma, X.~Xie, Y.~Huang, J.~Wang, and Y.~Liu,
  ``{FakeSpotter: A Simple yet Robust Baseline for Spotting AI-Synthesized Fake
  Faces},'' \emph{International Joint Conference on Artificial Intelligence
  (IJCAI)}, 2020.

\bibitem{goodfellow2014explaining}
I.~J. Goodfellow, J.~Shlens, and C.~Szegedy, ``Explaining and harnessing
  adversarial examples,'' in \emph{International Conference on Learning
  Representations}, 2015.

\bibitem{moosavi2016deepfool}
S.-M. Moosavi-Dezfooli, A.~Fawzi, and P.~Frossard, ``Deepfool: a simple and
  accurate method to fool deep neural networks,'' in \emph{Proceedings of the
  IEEE Conference on Computer Vision and Pattern Recognition}, 2016, pp.
  2574--2582.

\bibitem{su2019one}
J.~Su, D.~V. Vargas, and K.~Sakurai, ``One pixel attack for fooling deep neural
  networks,'' \emph{IEEE Transactions on Evolutionary Computation}, 2019.

\bibitem{guo2020spark}
Q.~Guo, X.~Xie, F.~Juefei-Xu, L.~Ma, Z.~Li, W.~Xue, W.~Feng, and Y.~Liu,
  ``Spark: Spatial-aware online incremental attack against visual tracking,''
  in \emph{Proceedings of the European Conference on Computer Vision (ECCV)},
  2020.

\bibitem{neurips20_abba}
Q.~Guo, F.~Juefei-Xu, X.~Xie, L.~Ma, J.~Wang, B.~Yu, W.~Feng, and Y.~Liu,
  ``{Watch out! Motion is Blurring the Vision of Your Deep Neural Networks},''
  in \emph{Advances in Neural Information Processing Systems (NeurIPS)}, 2020.

\bibitem{acmmm20_amora}
R.~Wang, F.~Juefei-Xu, Q.~Guo, Y.~Huang, X.~Xie, L.~Ma, and Y.~Liu, ``{Amora:
  Black-box Adversarial Morphing Attack},'' in \emph{Proceedings of the ACM
  International Conference on Multimedia (ACM MM)}, 2020.

\bibitem{tian2021ava}
B.~Tian, F.~Juefei-Xu, Q.~Guo, X.~Xie, X.~Li, and Y.~Liu, ``Ava: Adversarial
  vignetting attack against visual recognition,'' in \emph{Proceedings of the
  International Joint Conference on Artificial Intelligence (IJCAI)}, 2021.

\bibitem{Bako2017TOG}
S.~Bako, T.~Vogels, B.~Mcwilliams, M.~Meyer, J.~Nov\'{a}K, A.~Harvill, P.~Sen,
  T.~Derose, and F.~Rousselle, ``Kernel-predicting convolutional networks for
  denoising monte carlo renderings,'' \emph{ACM Trans. Graph.}, vol.~36, no.~4,
  Jul. 2017.

\bibitem{rahman2007video}
S.~M. Rahman, M.~O. Ahmad, and M.~Swamy, ``Video denoising based on inter-frame
  statistical modeling of wavelet coefficients,'' \emph{IEEE Transactions on
  Circuits and Systems for Video Technology}, vol.~17, no.~2, pp. 187--198,
  2007.

\bibitem{luisier2010sure}
F.~Luisier, T.~Blu, and M.~Unser, ``Sure-let for orthonormal wavelet-domain
  video denoising,'' \emph{IEEE Transactions on Circuits and Systems for Video
  Technology}, vol.~20, no.~6, pp. 913--919, 2010.

\bibitem{jiang2016srlsp}
J.~Jiang, C.~Chen, J.~Ma, Z.~Wang, Z.~Wang, and R.~Hu, ``Srlsp: A face image
  super-resolution algorithm using smooth regression with local structure
  prior,'' \emph{IEEE Transactions on Multimedia}, vol.~19, no.~1, pp. 27--40,
  2016.

\bibitem{ding2015robust}
C.~Ding and D.~Tao, ``Robust face recognition via multimodal deep face
  representation,'' \emph{IEEE Transactions on Multimedia}, vol.~17, no.~11,
  pp. 2049--2058, 2015.

\bibitem{perona1990scale}
P.~Perona and J.~Malik, ``Scale-space and edge detection using anisotropic
  diffusion,'' \emph{IEEE Transactions on pattern analysis and machine
  intelligence}, vol.~12, no.~7, pp. 629--639, 1990.

\bibitem{rudin1992nonlinear}
L.~I. Rudin, S.~Osher, and E.~Fatemi, ``Nonlinear total variation based noise
  removal algorithms,'' \emph{Physica D: nonlinear phenomena}, vol.~60, no.
  1-4, pp. 259--268, 1992.

\bibitem{chen2011adaptive}
M.~Chen, M.~Xu, and P.~Franti, ``Adaptive context-tree-based statistical
  filtering for raster map image denoising,'' \emph{IEEE Transactions on
  Multimedia}, vol.~13, no.~6, pp. 1195--1207, 2011.

\bibitem{dabov2007image}
K.~Dabov, A.~Foi, V.~Katkovnik, and K.~Egiazarian, ``Image denoising by sparse
  3-d transform-domain collaborative filtering,'' \emph{IEEE Transactions on
  Image Processing}, vol.~16, no.~8, pp. 2080--2095, 2007.

\bibitem{guo2015efficient}
Q.~Guo, C.~Zhang, Y.~Zhang, and H.~Liu, ``An efficient svd-based method for
  image denoising,'' \emph{IEEE Transactions on Circuits and Systems for Video
  Technology}, vol.~26, no.~5, pp. 868--880, 2015.

\bibitem{maggioni2011video}
M.~Maggioni, G.~Boracchi, A.~Foi, and K.~Egiazarian, ``Video denoising using
  separable 4d nonlocal spatiotemporal transforms,'' in \emph{Image Processing:
  Algorithms and Systems IX}, vol. 7870.\hskip 1em plus 0.5em minus 0.4em\relax
  International Society for Optics and Photonics, 2011, p. 787003.

\bibitem{heide2014flexisp}
F.~Heide, M.~Steinberger, Y.-T. Tsai, M.~Rouf, D.~Paj{\k{a}}k, D.~Reddy,
  O.~Gallo, J.~Liu, W.~Heidrich, K.~Egiazarian \emph{et~al.}, ``Flexisp: A
  flexible camera image processing framework,'' \emph{ACM Transactions on
  Graphics (TOG)}, vol.~33, no.~6, pp. 1--13, 2014.

\bibitem{heide2016proximal}
F.~Heide, S.~Diamond, M.~Nie{\ss}ner, J.~Ragan-Kelley, W.~Heidrich, and
  G.~Wetzstein, ``Proximal: Efficient image optimization using proximal
  algorithms,'' \emph{ACM Transactions on Graphics (TOG)}, vol.~35, no.~4, pp.
  1--15, 2016.

\bibitem{cho2018gradient}
S.~I. Cho and S.-J. Kang, ``Gradient prior-aided cnn denoiser with separable
  convolution-based optimization of feature dimension,'' \emph{IEEE
  Transactions on Multimedia}, vol.~21, no.~2, pp. 484--493, 2018.

\bibitem{szegedy2016rethinking}
C.~Szegedy, V.~Vanhoucke, S.~Ioffe, J.~Shlens, and Z.~Wojna, ``Rethinking the
  inception architecture for computer vision,'' in \emph{Proceedings of the
  IEEE Conference on Computer Vision and Pattern Recognition}, 2016, pp.
  2818--2826.

\bibitem{huang2017densely}
G.~Huang, Z.~Liu, L.~Van Der~Maaten, and K.~Q. Weinberger, ``Densely connected
  convolutional networks,'' in \emph{Proceedings of the IEEE Conference on
  Computer Vision and Rattern Recognition}, 2017, pp. 4700--4708.

\bibitem{He_2017_ICCV}
K.~He, G.~Gkioxari, P.~Dollar, and R.~Girshick, ``Mask r-cnn,'' in \emph{IEEE
  International Conference on Computer Vision}, 2017.

\bibitem{ChenPKMY18}
L.-C. Chen, G.~Papandreou, I.~Kokkinos, K.~Murphy, and A.~L. Yuille, ``Deeplab:
  Semantic image segmentation with deep convolutional nets, atrous convolution,
  and fully connected crfs.'' \emph{IEEE Transactions on Pattern Analysis and
  Machine Intelligence}, vol.~40, no.~4, pp. 834--848, 2018.

\bibitem{zhang2017beyond}
K.~Zhang, W.~Zuo, Y.~Chen, D.~Meng, and L.~Zhang, ``Beyond a gaussian denoiser:
  Residual learning of deep cnn for image denoising,'' \emph{IEEE Transactions
  on Image Processing}, vol.~26, no.~7, pp. 3142--3155, 2017.

\bibitem{chen2016deep}
X.~Chen, L.~Song, and X.~Yang, ``Deep rnns for video denoising,'' in
  \emph{Applications of Digital Image Processing XXXIX}, vol. 9971.\hskip 1em
  plus 0.5em minus 0.4em\relax International Society for Optics and Photonics,
  2016, p. 99711T.

\bibitem{su2017deep}
S.~Su, M.~Delbracio, J.~Wang, G.~Sapiro, W.~Heidrich, and O.~Wang, ``Deep video
  deblurring for hand-held cameras,'' in \emph{Proceedings of the IEEE
  Conference on Computer Vision and Pattern Recognition}, 2017, pp. 1279--1288.

\bibitem{liang2020raw}
C.-H. Liang, Y.-A. Chen, Y.-C. Liu, and W.~Hsu, ``Raw image deblurring,''
  \emph{IEEE Transactions on Multimedia}, 2020.

\bibitem{gharbi2016deep}
M.~Gharbi, G.~Chaurasia, S.~Paris, and F.~Durand, ``Deep joint demosaicking and
  denoising,'' \emph{ACM Transactions on Graphics (TOG)}, vol.~35, no.~6, pp.
  1--12, 2016.

\bibitem{tao2017detail}
X.~Tao, H.~Gao, R.~Liao, J.~Wang, and J.~Jia, ``Detail-revealing deep video
  super-resolution,'' in \emph{Proceedings of the IEEE International Conference
  on Computer Vision}, 2017, pp. 4472--4480.

\bibitem{ackerman2017drive}
E.~Ackerman, ``How drive. ai is mastering autonomous driving with deep
  learning,'' \emph{IEEE Spectrum Magazine}, vol.~1, 2017.

\bibitem{rao2018deep}
Q.~Rao and J.~Frtunikj, ``Deep learning for self-driving cars: chances and
  challenges,'' in \emph{Proceedings of the 1st International Workshop on
  Software Engineering for AI in Autonomous Systems}, 2018, pp. 35--38.

\bibitem{najafabadi2015deep}
M.~M. Najafabadi, F.~Villanustre, T.~M. Khoshgoftaar, N.~Seliya, R.~Wald, and
  E.~Muharemagic, ``Deep learning applications and challenges in big data
  analytics,'' \emph{Journal of Big Data}, vol.~2, no.~1, p.~1, 2015.

\bibitem{papernot2016towards}
N.~Papernot, P.~McDaniel, A.~Sinha, and M.~Wellman, ``Towards the science of
  security and privacy in machine learning,'' \emph{arXiv preprint
  arXiv:1611.03814}, 2016.

\bibitem{hinton2012deep}
G.~Hinton, L.~Deng, D.~Yu, G.~E. Dahl, A.-r. Mohamed, N.~Jaitly, A.~Senior,
  V.~Vanhoucke, P.~Nguyen, T.~N. Sainath \emph{et~al.}, ``Deep neural networks
  for acoustic modeling in speech recognition: The shared views of four
  research groups,'' \emph{IEEE Signal Processing Magazine}, vol.~29, no.~6,
  pp. 82--97, 2012.

\bibitem{szegedy2013intriguing}
C.~Szegedy, W.~Zaremba, I.~Sutskever, J.~Bruna, D.~Erhan, I.~Goodfellow, and
  R.~Fergus, ``Intriguing properties of neural networks,'' in
  \emph{International Conference on Learning Representations}, 2014.

\bibitem{BIM_2016_ICLRW}
A.~Kurakin, I.~Goodfellow, and S.~Bengio, ``Adversarial examples in the
  physical world,'' \emph{International Conference on Learning Representations
  (Workshop)}, 2017.

\bibitem{dong2018boosting}
Y.~Dong, F.~Liao, T.~Pang, H.~Su, J.~Zhu, X.~Hu, and J.~Li, ``Boosting
  adversarial attacks with momentum,'' in \emph{Proceedings of the IEEE
  Conference on Computer Vision and Pattern Recognition}, 2018, pp. 9185--9193.

\bibitem{xie2019improving}
C.~Xie, Z.~Zhang, Y.~Zhou, S.~Bai, J.~Wang, Z.~Ren, and A.~L. Yuille,
  ``Improving transferability of adversarial examples with input diversity,''
  in \emph{Proceedings of the IEEE Conference on Computer Vision and Pattern
  Recognition}, 2019, pp. 2730--2739.

\bibitem{Papernot2016TheLO}
N.~Papernot, P.~D. McDaniel, S.~Jha, M.~Fredrikson, Z.~B. Celik, and A.~Swami,
  ``The limitations of deep learning in adversarial settings,'' \emph{IEEE
  European Symposium on Security and Privacy (EuroS P)}, pp. 372--387, 2016.

\bibitem{CW_2017_SSP}
N.~{Carlini} and D.~{Wagner}, ``Towards evaluating the robustness of neural
  networks,'' in \emph{2017 IEEE Symposium on Security and Privacy (SP)}, 2017,
  pp. 39--57.

\bibitem{He2019CVPR}
J.~He, S.~Zhang, M.~Yang, Y.~Shan, and T.~Huang, ``Bi-directional cascade
  network for perceptual edge detection,'' in \emph{Proceedings of the IEEE
  Conference on Computer Vision and Pattern Recognition}, 2019, pp. 3828--3837.

\bibitem{fu2021auto}
L.~Fu, C.~Zhou, Q.~Guo, F.~Juefei-Xu, H.~Yu, W.~Feng, Y.~Liu, and S.~Wang,
  ``Auto-exposure fusion for single-image shadow removal,'' \emph{arXiv
  preprint arXiv:2103.01255}, 2021.

\bibitem{guo2021efficientderain}
Q.~Guo, J.~Sun, F.~Juefei-Xu, L.~Ma, X.~Xie, W.~Feng, and Y.~Liu,
  ``Efficientderain: Learning pixel-wise dilation filtering for high-efficiency
  single-image deraining,'' \emph{AAAI conference on artificial intelligence},
  2021.

\bibitem{ghosh2017artifact}
S.~Ghosh and K.~N. Chaudhury, ``Artifact reduction for separable nonlocal
  means,'' \emph{Journal of Electronic Imaging}, vol.~26, no.~6, p. 063012,
  2017.

\bibitem{ghosh2017pruned}
S.~Ghosh, A.~K. Mandal, and K.~N. Chaudhury, ``Pruned non-local means,''
  \emph{IET Image Processing}, vol.~11, no.~5, pp. 317--323, 2017.

\bibitem{kurakin2018adversarial}
A.~Kurakin, I.~Goodfellow, S.~Bengio, Y.~Dong, F.~Liao, M.~Liang, T.~Pang,
  J.~Zhu, X.~Hu, C.~Xie \emph{et~al.}, ``Adversarial attacks and defences
  competition,'' in \emph{The NIPS'17 Competition: Building Intelligent
  Systems}.\hskip 1em plus 0.5em minus 0.4em\relax Springer, 2018, pp.
  195--231.

\bibitem{cvpr2021-aic-vi}
``Adversarial attacks on ml defense models at cvpr 2021 workshop,'' Website,
  2021,
  \url{https://tianchi.aliyun.com/competition/entrance/531847/introduction?lang=en-us}.

\bibitem{dong2020benchmarking}
Y.~Dong, Q.-A. Fu, X.~Yang, T.~Pang, H.~Su, Z.~Xiao, and J.~Zhu, ``Benchmarking
  adversarial robustness on image classification,'' in \emph{Proceedings of the
  IEEE/CVF Conference on Computer Vision and Pattern Recognition (CVPR)}, 2020,
  pp. 321--331.

\bibitem{hendrycks2019robustness}
D.~Hendrycks and T.~Dietterich, ``Benchmarking neural network robustness to
  common corruptions and perturbations,'' \emph{Proceedings of the
  International Conference on Learning Representations}, 2019.

\bibitem{szegedy2017inception}
C.~Szegedy, S.~Ioffe, V.~Vanhoucke, and A.~A. Alemi, ``Inception-v4,
  inception-resnet and the impact of residual connections on learning,'' in
  \emph{Thirty-first AAAI conference on artificial intelligence}, 2017.

\bibitem{wang2004image}
Z.~Wang, A.~C. Bovik, H.~R. Sheikh, and E.~P. Simoncelli, ``Image quality
  assessment: from error visibility to structural similarity,'' \emph{IEEE
  Transactions on Image Processing}, vol.~13, no.~4, pp. 600--612, 2004.

\bibitem{mittal2012no}
A.~Mittal, A.~K. Moorthy, and A.~C. Bovik, ``No-reference image quality
  assessment in the spatial domain,'' \emph{IEEE Transactions on Image
  Processing}, vol.~21, no.~12, pp. 4695--4708, 2012.

\bibitem{croce2019sparse}
F.~Croce and M.~Hein, ``Sparse and imperceivable adversarial attacks,'' in
  \emph{Proceedings of the IEEE International Conference on Computer Vision},
  2019, pp. 4724--4732.

\bibitem{canny1986computational}
J.~Canny, ``A computational approach to edge detection,'' \emph{IEEE
  Transactions on pattern analysis and machine intelligence}, no.~6, pp.
  679--698, 1986.

\bibitem{wong2020fast}
E.~Wong, L.~Rice, and J.~Z. Kolter, ``Fast is better than free: Revisiting
  adversarial training,'' in \emph{International Conference on Learning
  Representations}, 2020.

\bibitem{tan2019efficientnet}
M.~Tan and Q.~Le, ``Efficientnet: Rethinking model scaling for convolutional
  neural networks,'' in \emph{International Conference on Machine
  Learning}.\hskip 1em plus 0.5em minus 0.4em\relax PMLR, 2019, pp. 6105--6114.

\bibitem{iccv21_advmot}
Q.~Guo, Z.~Cheng, F.~Juefei-Xu, L.~Ma, X.~Xie, Y.~Liu, and J.~Zhao, ``{Learning
  to Adversarially Blur Visual Object Tracking},'' in \emph{Proceedings of the
  IEEE International Conference on Computer Vision (ICCV)}.\hskip 1em plus
  0.5em minus 0.4em\relax IEEE, October 2021.

\bibitem{zhai2020s}
L.~Zhai, F.~Juefei-Xu, Q.~Guo, X.~Xie, L.~Ma, W.~Feng, S.~Qin, and Y.~Liu,
  ``It's raining cats or dogs? adversarial rain attack on dnn perception,''
  \emph{arXiv preprint arXiv:2009.09205}, 2020.

\bibitem{gao2021advhaze}
R.~Gao, Q.~Guo, F.~Juefei-Xu, H.~Yu, and W.~Feng, ``Advhaze: Adversarial haze
  attack,'' \emph{arXiv preprint arXiv:2104.13673}, 2021.

\end{thebibliography}

\vspace{-3em}
\begin{IEEEbiography}[{\includegraphics[width=1in,height=1.25in,clip,keepaspectratio]{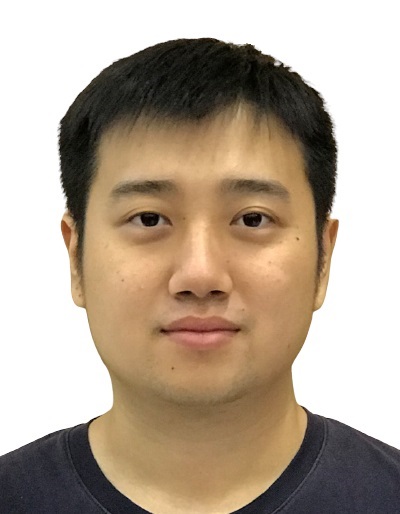}}]{Yupeng Cheng} received his B.S. degree and M. E. degree from the School of Computer Science and Technology, Tianjin University, China. He is currently pursuing Ph. D. degree in the Nanyang Technological University, Singapore, from 2016. His research interests include computer vision, AI security, and image processing.
\end{IEEEbiography}

\vspace{-3em}
\begin{IEEEbiography}[{\includegraphics[width=1in,height=1.25in,clip,keepaspectratio]{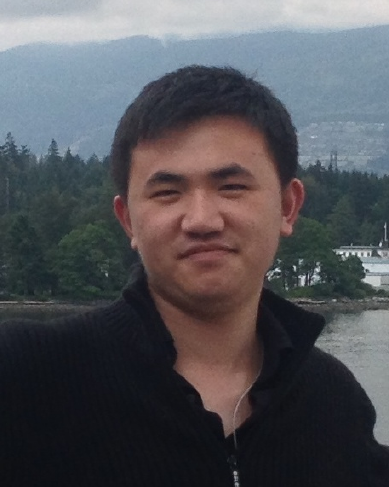}}]{Qing Guo} received his B.S. degree in Electronic and Information Engineering from the North China Institute of Aerospace Engineering in 2011, M.E. degree in computer application technology from the College of Computer and Information Technology, China Three Gorges University in 2014, and the Ph.D. degree in computer application technology from the School of Computer Science and Technology, Tianjin University, China. He was a research fellow with the Nanyang Technological University, Singapore, from Dec. 2019 to Sep. 2020. He is currently a Wallenberg-NTU Presidential Postdoctoral Fellow with the Nanyang Technological University, Singapore. His research interests include computer vision, AI security, and image processing. He is a member of IEEE.
\end{IEEEbiography}

\vspace{-3em}
\begin{IEEEbiography}[{\includegraphics[width=1in,height=1.25in,clip]{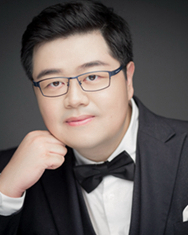}}]{Felix Juefei-Xu} received the Ph.D. degree in Electrical and Computer Engineering from Carnegie Mellon University (CMU), Pittsburgh, PA, USA. Prior to that, he received the M.S. degree in Electrical and Computer Engineering and the M.S degree in Machine Learning from CMU, and the B.S. degree in Electronic Engineering from Shanghai Jiao Tong University (SJTU), Shanghai, China. Currently, he is a Research Scientist with Alibaba Group, Sunnyvale, CA, USA, with research focus on a fuller understanding of deep learning where he is actively exploring new methods in deep learning that are statistically efficient and adversarially robust. He also has broader interests in pattern recognition, computer vision, machine learning, optimization, statistics, compressive sensing, and image processing. He is the recipient of multiple best/distinguished paper awards, including IJCB'11, BTAS'15-16, ASE'18, and ACCV'18.
\end{IEEEbiography}

\vspace{-3em}
\begin{IEEEbiography}[{\includegraphics[width=1in,height=1.25in,clip,keepaspectratio]{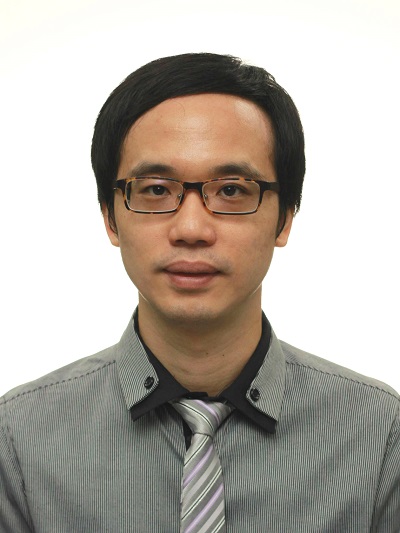}}]{Shang-Wei Lin}
LIN Shang-Wei, Ph.D., received his B.S. degree in Information Management from the National Chung Cheng University, Chiayi, Taiwan, in 2003 and received his Ph.D. degree in Computer Science and Information Engineering from the National Chung Cheng University, Chiayi, Taiwan, in 2010. From September 2003 to July 2010, he was a teaching and research assistant in the Department of Computer Science and Information Engineering at the National Chung Cheng University. In 2011, he was a postdoctoral researcher at School of Computing, National University of Singapore (NUS). From 2012 to November 2014, he was a research scientist at Temasek Laboratories in National University of Singapore (NUS). He was also the principal investigator of the seed project of Temasek Laboratories in 2013. From December 2014 to April 2015, he was a postdoctoral research fellow in Singapore University of Technology and Design (SUTD). He has joined School of Computer Science and Engineering, Nanyang Technological University (NTU) as Assistant Professor in April 2015. His research interests include formal verification, formal synthesis, embedded system design, cyber-physical systems, security systems, multi-core programming, and component-based object-oriented application frameworks for real-time embedded systems. Recently, he is working on applying formal methods on smart contract verification.
\end{IEEEbiography}

\vspace{-3em}
\begin{IEEEbiography}[{\includegraphics[width=1in,height=1.25in,clip,keepaspectratio]{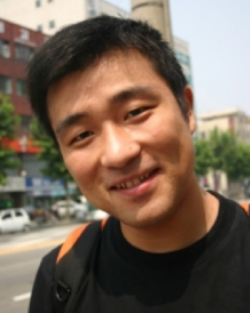}}]{Wei Feng}
(M' 06) received the PhD degree in computer science from City University of Hong Kong in 2008. From 2008 to 2010, he was a research fellow at the Chinese University of Hong Kong and City University of Hong Kong. He is now a full Professor at the School of Computer Science and Technology, College of Computing and Intelligence, Tianjin University, China. His major research interests are active robotic vision and visual intelligence, specifically including active camera relocalization and lighting recurrence, general Markov Random Fields modeling, energy minimization, active 3D scene perception, SLAM, video analysis, and generic pattern recognition. Recently, he focuses on solving preventive conservation problems of cultural heritages via computer vision and machine learning. He is the Associate Editor of Neurocomputing and Journal of Ambient Intelligence and Humanized Computing.
\end{IEEEbiography}

\vspace{-3em}
\begin{IEEEbiography}[{\includegraphics[width=1in,height=1.25in,clip,keepaspectratio]{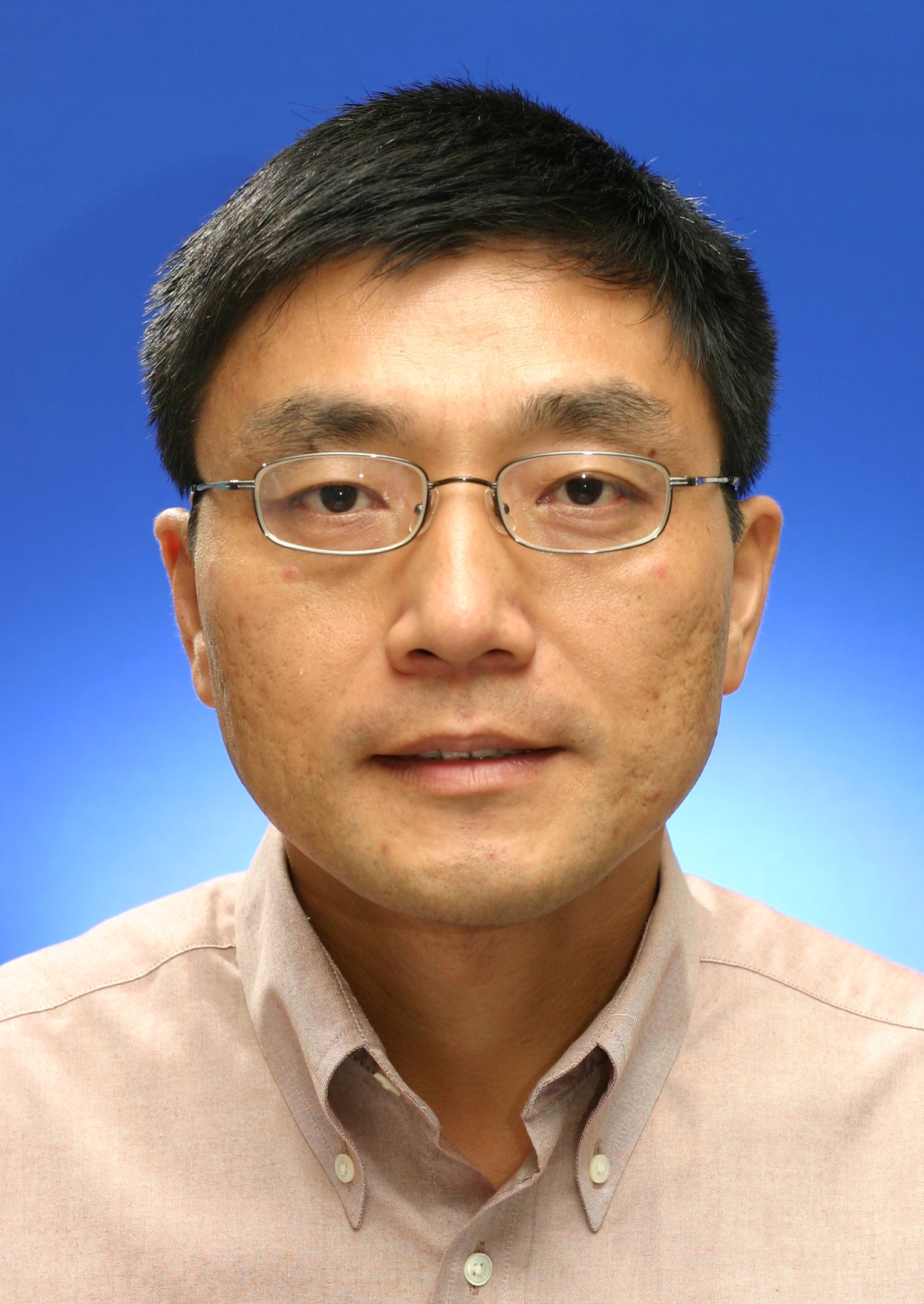}}]{Weisi Lin}
Weisi Lin (M’92-SM’98-F’16) is a Professor and the Associate Chair (Research) in the School of Computer Science and Engineering, Nanyang Technological University, Singapore. His research interests include intelligent image processing, perceptual signal modeling, video compression, and multimedia communication. He is a Chartered Engineer and a fellow of the IET. He was the Technical Program Chair of the IEEE ICME 2013, PCM 2012, QoMEX 2014, and the IEEE VCIP 2017. He has been a Keynote/Invited/Panelist/Tutorial Speaker at over 30 international conferences and was a Distinguished Lecturer of the IEEE Circuits and Systems Society from 2016 to 2017 and the AsiaPacific Signal and Information Processing Association (APSIPA) from 2012 to 2013. He has been an Associate Editor of the IEEE Trans. Image Process., the IEEE Trans. Circuits Syst. Video Technol., the IEEE Trans. Multimedia, and the IEEE Signal Process. Lett. 
\end{IEEEbiography}

\vspace{-3em}
\begin{IEEEbiography}[{\includegraphics[width=1in,height=1.25in,clip,keepaspectratio]{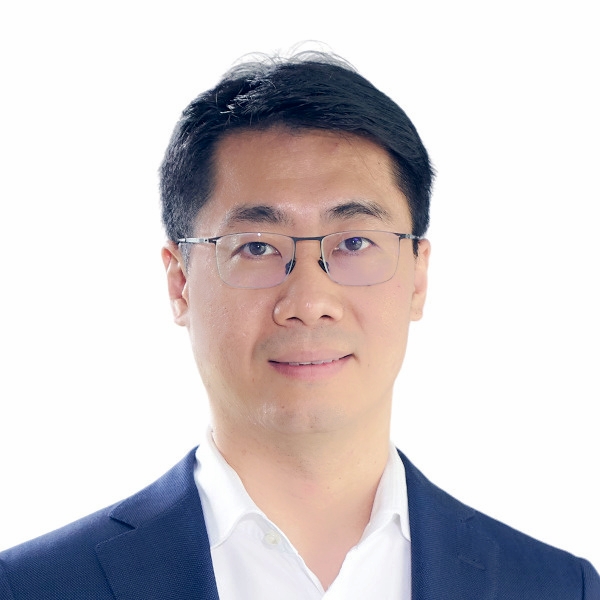}}]{Yang Liu}
graduated in 2005 with a Bachelor of
Computing (Honours) in the National University
of Singapore (NUS). In 2010, he obtained his
Ph.D. and started his post-doctoral work in NUS
and MIT. In 2012, he joined Nanyang Technological University (NTU), and currently is a full
professor and Director of the cybersecurity lab in
NTU.
Dr. Liu specializes in software engineering, cybersecurity and artificial intelligence. His research has bridged the gap between the theory and practical usage of program analysis, data analysis and AI to evaluate the design and implementation of software for high assurance and security. By now, he has more than 400 publications in top tier conferences and journals. He has received a number of prestigious awards including MSRA Fellowship, TRF Fellowship, Nanyang Assistant Professor, Tan Chin Tuan Fellowship, Nanyang Research Award 2019, ACM Distinguished Speaker, NRF Investigatorship, and 15 best paper awards and one most influence system award in top software engineering conferences like ASE, FSE and ICSE.
\end{IEEEbiography}

\end{document}